\documentclass{article}

\PassOptionsToPackage{dvipsnames,table}{xcolor}

\usepackage{iclr2027_conference,times}

\usepackage{amsmath,amsfonts,bm}

\def\eqref#1{equation~\ref{#1}}

\def\1{\bm{1}}

\def\vzero{{\bm{0}}}

\def\vb{{\bm{b}}}

\def\vr{{\bm{r}}}

\def\vu{{\bm{u}}}

\def\vw{{\bm{w}}}
\def\vx{{\bm{x}}}
\def\vy{{\bm{y}}}
\def\vz{{\bm{z}}}

\def\mA{{\bm{A}}}
\def\mB{{\bm{B}}}

\def\mG{{\bm{G}}}
\def\mH{{\bm{H}}}
\def\mI{{\bm{I}}}

\def\mK{{\bm{K}}}
\def\mL{{\bm{L}}}
\def\mM{{\bm{M}}}

\def\mP{{\bm{P}}}
\def\mQ{{\bm{Q}}}
\def\mR{{\bm{R}}}
\def\mS{{\bm{S}}}
\def\mT{{\bm{T}}}

\def\mV{{\bm{V}}}
\def\mW{{\bm{W}}}

\def\mPhi{{\bm{\Phi}}}
\def\mLambda{{\bm{\Lambda}}}
\def\mSigma{{\bm{\Sigma}}}

\DeclareMathAlphabet{\mathsfit}{\encodingdefault}{\sfdefault}{m}{sl}
\SetMathAlphabet{\mathsfit}{bold}{\encodingdefault}{\sfdefault}{bx}{n}

\newcommand{\E}{\mathbb{E}}

\let\originalleft\left
\let\originalright\right
\renewcommand{\left}{\mathopen{}\mathclose\bgroup\originalleft}
\renewcommand{\right}{\aftergroup\egroup\originalright}

\let\originalpartial\partial
\renewcommand{\partial}{\mathop{}\!\mathrm{\originalpartial}}

\newcommand{\T}{^\mathrm{T}}

\usepackage{hyperref}
\usepackage{url}

\usepackage{amssymb}
\usepackage{mathtools}
\usepackage{amsthm}
\usepackage{graphicx}
\usepackage{booktabs}
\usepackage{caption}
\usepackage{subcaption}
\usepackage{enumitem}
\usepackage{xcolor}
\usepackage{float}
\usepackage{wrapfig}
\usepackage{placeins}
\usepackage{multirow}
\usepackage{makecell}

\newlength\itemparskip
\setlist{nosep, leftmargin=1em, itemsep=0pt, parsep=0pt, before={\parskip=\itemparskip}, after=\vspace{-\itemparskip}}%

\newtheoremstyle{custom}
{\parskip} %
{0}
{} %
{} %
{\bfseries} %
{.} %
{.5em} %
{} %
\theoremstyle{custom}
\newtheorem{theorem}{Theorem}
\newtheorem{proposition}[theorem]{Proposition}

\theoremstyle{definition}

\newtheorem{assumption}[theorem]{Assumption}
\theoremstyle{remark}
\newtheorem{remark}[theorem]{Remark}

\iclrfinalcopy
\arxiv

\begin{document}

\title{Learning Anisotropic Prediction Targets for Flow Models}

\author{%
  \vspace{1ex}
  \textbf{Rayhan Zirvi}$^{1,2*}$\quad
  \textbf{Hansheng Chen}$^{1*}$\quad
  \textbf{Gordon Wetzstein}$^{1}$\quad
  \textbf{Leonidas Guibas}$^{1}$ \\
  \vspace{1ex}
  $^{1}$Stanford University\quad
  $^{2}$California Institute of Technology\quad
  $^{*}$Equal contribution \\
  \vspace{-1ex}
  \url{https://hanshengchen.com/anisoflow}
}

\maketitle

\begin{figure}[H]\centering
\begin{subfigure}[t]{0.623\linewidth}
\vspace{0pt}\centering
\includegraphics[width=\linewidth]{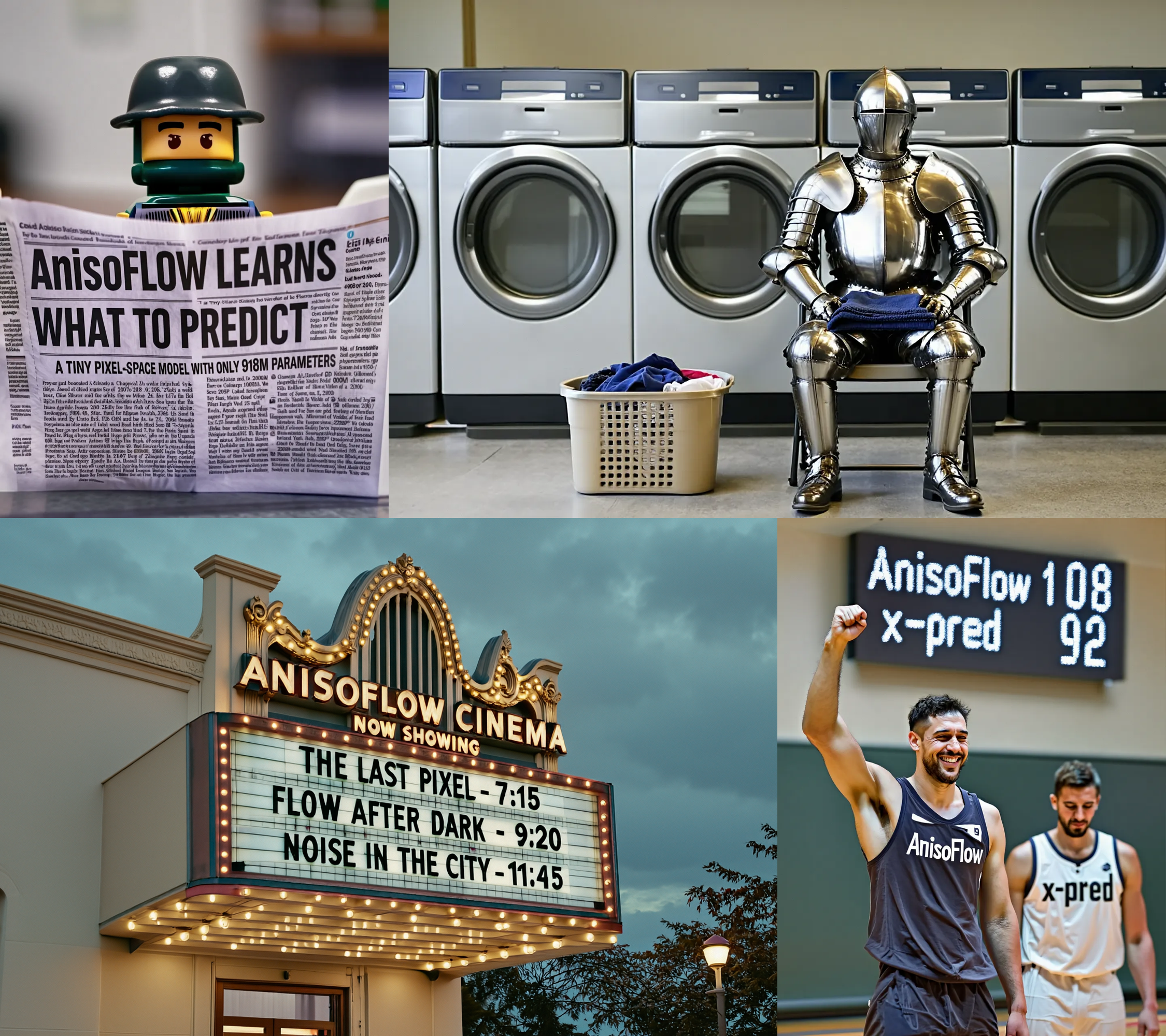}
\caption{Text-to-image generations in pixel space}
\label{fig:teaser-gen}
\end{subfigure}
\hfill
\begin{subfigure}[t]{0.35\linewidth}
\vspace{0pt}\centering
\includegraphics[width=\linewidth]{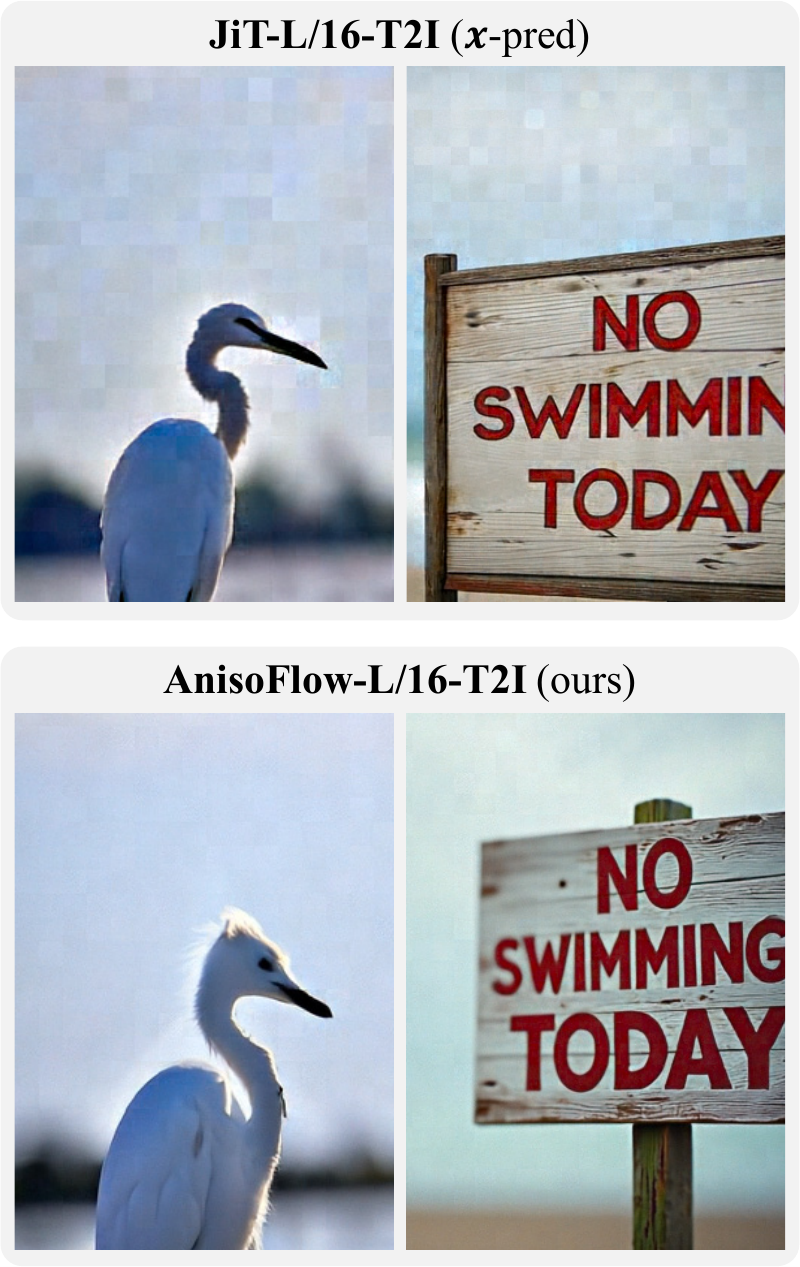}
\caption{FP8 blockwise inference}
\label{fig:teaser-artifact}
\end{subfigure}
\caption{
\textbf{(a)}~High-resolution images generated by AnisoFlow, a pixel-space diffusion transformer with only 918M parameters trained from scratch, using a learned anisotropic prediction target.
\textbf{(b)}~By learning a well-conditioned prediction target, AnisoFlow removes the
patch-seam artifacts produced by the $\vx$-prediction baseline under FP8 blockwise (W8A8) quantization~\citep{deepseekai}.}
\label{fig:teaser}
\end{figure}

\begin{abstract}
    Flow-based generative models are trained on a hand-chosen prediction target, such as the velocity, the noise, or the clean data. This choice strongly affects training and sample quality. We introduce \emph{Anisotropic Flow Models} (AnisoFlow), which replaces a fixed prediction target with a fully learnable one. AnisoFlow learns to scale the noise component of the target anisotropically, assigning a separate nonnegative weight to each direction in a jointly learned basis. A closed-form map recovers the velocity from the network's prediction, leaving the architecture, flow-matching loss, and sampler unchanged. Since recovery is exact, training optimizes the target purely to make it easier for the network to predict. We theoretically analyze the resulting prediction error and predict that the preferred noise weight in each direction scales near-linearly with data variance, in agreement with the weights learned empirically in pixel space. AnisoFlow improves FID from $1.90$ to $1.55$ over the JiT baseline on ImageNet $256{\times}256$, by learning only the prediction target, without representation alignment or architectural changes. In pixel-space text-to-image generation, AnisoFlow improves image quality and, through better-conditioned velocity recovery, largely removes patch-seam artifacts under reduced-precision inference.\ifarxiv\else{} Code and models will be released.\fi
\end{abstract}

\section{Introduction}

Modern image and video generators use diffusion or flow objectives \citep{ho2020ddpm, song2021sde, flowmatching, rectifiedflow, albergo2023interpolants} to train scalable denoising networks \citep{dit, brooks2024sora} to predict a target along the noising path, such as the noise $\bm{\epsilon}$, the clean data $\vx$, or the velocity $\vu \coloneqq \bm{\epsilon} - \vx$. The choice of target is a central design decision: given the noisy input and timestep, these quantities carry the same information, yet they define distinct regression problems with substantially different training behavior and sample quality \citep{edm, salimans2022progressivedistillation}. Existing hand-designed choices occupy a small corner of a much larger design space: velocity in latent models \citep{esser2024sd3, bfl2025flux2, hacohen2024ltxvideo, kong2025hunyuanvideo, wang2025wan} or clean data in recent pixel models \citep{jit, guo2026pixelu}, with methods also exploring interpolation between standard targets \citep{kdiff} or masking part of the noise component \citep{asymflow}.

We introduce \emph{Anisotropic Flow Models} (AnisoFlow), which replaces this hand-designed choice with a fully learnable one. AnisoFlow learns to scale the noise component of the target anisotropically.
It defines the target $\vu_W \coloneqq \mW\bm{\epsilon} - \vx$, where $\mW$ is a positive semidefinite (PSD) matrix that applies a separate learnable nonnegative weight $w_i$ to the noise along each direction of a jointly learned orthogonal basis. Up to sign and overall scale, the target family includes noise prediction at $w_i \to +\infty$, clean-data prediction at $w_i=0$, velocity prediction at $w_i=1$, shared scalar interpolations at $w_i\equiv w$, and binary low-rank masks at $w_i\in\{0,1\}$. A closed-form map recovers the velocity from the network's prediction, leaving the architecture, flow-matching loss, and sampler unchanged.

AnisoFlow is particularly effective in pixel space, where patchification places many pixel coordinates within each transformer token, exposing directions with widely varying data variance \citep{jit, asymflow} and creating a setting where different directions can benefit from different prediction targets. The velocity target places unit noise weight in every direction, increasing the difficulty of prediction in low-variance directions. Clean-data prediction removes the noise component but can be harder in high-variance directions and makes velocity recovery ill-conditioned at low noise. AnisoFlow resolves this tradeoff by learning a separate noise weight for each direction.

We theoretically analyze which noise weights make prediction easiest for the network. We study the prediction error in the PCA basis, which diagonalizes the data covariance and closely matches the basis that AnisoFlow learns empirically. Under a local frozen-error model, we derive the preferred weight in each direction and show that it scales near-linearly with data variance. Consistent with this analysis, pixel-space models learn weights that follow the scaling $w_i \approx c\lambda_i$, where $\lambda_i$ is the data variance in direction $i$.

Empirically, AnisoFlow improves FID from $1.90$ to $1.55$ over the JiT-H/16 baseline on ImageNet $256{\times}256$ in pixel space, by learning only the prediction target, without representation alignment or architectural changes. In pixel-space text-to-image generation, AnisoFlow improves image quality and, through better-conditioned velocity recovery, largely eliminates patch-seam artifacts under reduced-precision inference. In latent-space models, AnisoFlow recovers weights close to the default velocity target in SiT \citep{ma2024sit} with SD-VAE \citep{ldm} and the default clean-data target in RAEv2 \citep{singh2026raev2}, achieving comparable performance.

\textbf{Our contributions} include (a)~AnisoFlow, a fully learnable prediction-target family that subsumes standard targets as special cases; (b)~a theoretical analysis relating the preferred noise weights to the network's prediction errors, explaining the observed near-linear scaling and approximate timestep invariance; and (c)~state-of-the-art pixel-space generation on ImageNet and improved quality and stability in text-to-image generation.

\section{Related Work}
\label{sec:related}

\textbf{Standard prediction parameterizations.}
The choice among noise ($\bm{\epsilon}$)-, velocity ($\vu$)-, and clean-data ($\vx$)-prediction determines how
the target combines the data and noise components and can substantially affect
training behavior and sample quality
\citep{edm, salimans2022progressivedistillation, kdiff}.
Noise- and velocity-prediction assign a nonzero noise
coefficient to every direction, which can make regression difficult in
high-dimensional pixel space
\citep{jit, asymflow, kdiff}. Clean-data prediction avoids predicting the
noise, but recovering the velocity from the predicted data
amplifies its error by a factor proportional to $1/\sigma_t$, which is
ill-conditioned at low noise
\citep{edm, salimans2022progressivedistillation, asymflow}.

\textbf{Interpolated targets.}
Isotropic interpolations mix data and noise with the same coefficient in every direction: EDM~\citep{edm} sets this coefficient analytically through a $\sigma_t$-dependent skip connection, and k-Diff \citep{kdiff} learns it as a single scalar. AsymFlow \citep{asymflow} is anisotropic but binary, combining velocity prediction in a low-rank subspace and data prediction in the orthogonal complement. Both k-Diff and AsymFlow lie within AnisoFlow's target family, as a shared scalar and a binary profile. Appendix~\ref{app:learned-weights} shows that a per-direction extension of EDM's choice is exactly what our analysis predicts when the network has no information beyond the timestep. Recent work also studies how prediction parameterizations interact with representation dimension, covariance conditioning, and low-variance directions \citep{fu2026jlt, gagneux2026training, zhang2026rit, wang2025analytical, wang2026rmt, ahamed2026preconditioned}.

\textbf{Rotation-based quantization.}
The rotate-then-scale form of AnisoFlow's recovery resembles rotation-based quantization~\citep{ashkboos2024quarot,liu2025spinquant}, which applies fixed or learned orthogonal transforms to weights and activations to suppress outliers before quantization. These methods operate layer-wise on a trained network, whereas AnisoFlow learns network-wide preconditioning through the prediction target, jointly with the network.

\section{Preliminaries}
\label{sec:prelim}

\textbf{Flow matching.}
Let $\vx\sim p_{\mathrm{data}}$ denote a data sample and let
$\bm{\epsilon}\sim\mathcal N(\vzero,\mI)$ denote Gaussian noise. We consider the
linear path
\begin{equation}
\vz_t = \alpha_t \vx + \sigma_t \bm{\epsilon},
\qquad
\alpha_t = 1-t,
\qquad
\sigma_t = t,
\qquad
t\in[0,1],
\end{equation}
so that $\vz_0=\vx$ and $\vz_1=\bm{\epsilon}$. Its sample velocity is
$\vu = \partial_t \vz_t = \bm{\epsilon}-\vx$.
A flow model $\vu_\theta(\vz_t,t)$ is trained by regressing onto this sample
velocity; under squared error, the optimal predictor is the conditional mean
$\E[\vu\mid\vz_t,t]$, which defines the ODE velocity field. The training
objective is
\begin{equation}
\mathcal L_{\mathrm{FM}}
=
\E_{\substack{t\sim p(t),\,\vx\sim p_{\mathrm{data}},\\
\bm{\epsilon}\sim\mathcal N(\vzero,\mI)}}
\left[
\left\|
\vu_\theta(\vz_t,t)-\vu
\right\|^2
\right].
\end{equation}
At sampling time, $\vz_1\sim\mathcal N(\vzero,\mI)$ is integrated from
$t=1$ to $t=0$ using the learned velocity field
\citep{flowmatching, rectifiedflow, albergo2023interpolants}.

\textbf{Prediction targets and recovery.}
The network need not predict the velocity $\vu$ directly; it may instead
predict the noise $\bm{\epsilon}$, the clean data $\vx$, or another linearly
related target. For any interior timestep $t\in(0,1)$, the relation
$\vz_t=\alpha_t\vx+\sigma_t\bm{\epsilon}$
allows either standard prediction to be converted back to velocity. Using
$\alpha_t+\sigma_t=1$, a noise prediction $\hat{\bm{\epsilon}}_\theta$ gives
$\hat{\vu}=(\hat{\bm{\epsilon}}_\theta-\vz_t)/\alpha_t$,
while a clean-data prediction $\hat{\vx}_\theta$ gives
$\hat{\vu}=(\vz_t-\hat{\vx}_\theta)/\sigma_t$.
More generally, any linearly related target that can be inverted together
with $\vz_t$ can be mapped to the velocity by a closed-form linear map.
After this recovery, the different prediction parameterizations use the same
velocity-based sampler.

\section{Anisotropic Flow Models}
\label{sec:method}

\begin{figure}[t]\centering
\begin{subfigure}[t]{0.7\linewidth}\centering
\includegraphics[width=\linewidth]{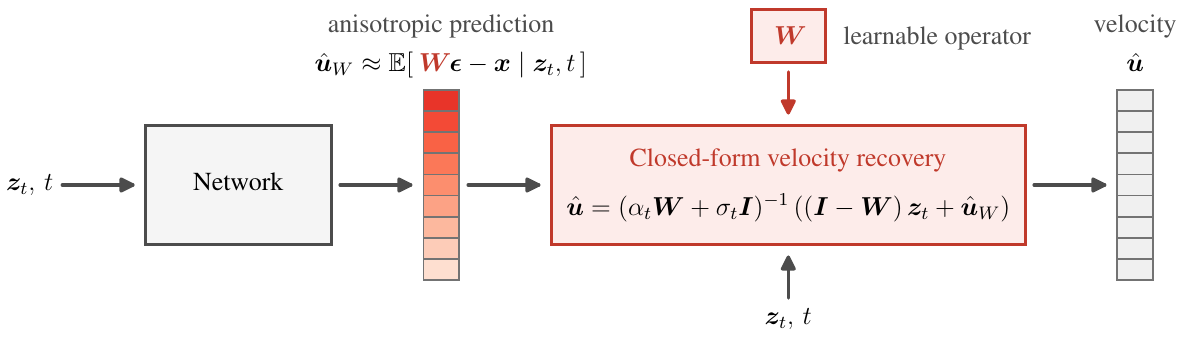}
\vspace{-4ex}
\caption{Prediction and exact recovery}\label{fig:overview-recovery}
\end{subfigure}\\[1.5ex]
\begin{subfigure}[t]{0.78\linewidth}\centering
\includegraphics[width=\linewidth]{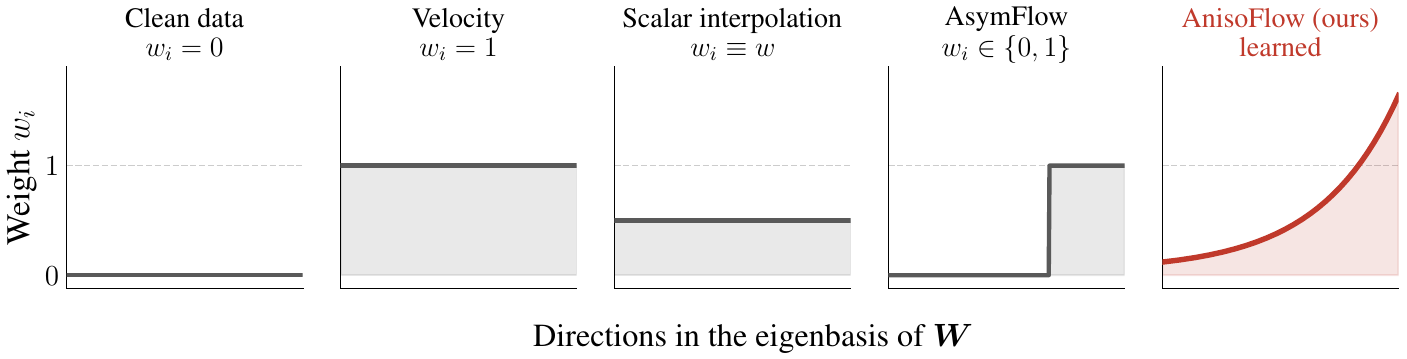}
\vspace{-3ex}
\caption{Target family}\label{fig:overview-family}
\end{subfigure}
\caption{\textbf{AnisoFlow overview.}
\textbf{(a)}~The network predicts the AnisoFlow target
$\vu_W=\mW\bm{\epsilon}-\vx$, where $\mW$ is a learned positive semidefinite
operator, and a closed-form map recovers the standard velocity.
\textbf{(b)}~In the eigenbasis of $\mW$, the target is parameterized separately in each
direction, retaining the clean-data term $-x_i$ and applying a learned
nonnegative weight $w_i$ to the noise term:
$u_{W,i}=w_i\epsilon_i-x_i$. Standard targets correspond to fixed weight
profiles. AnisoFlow instead learns a
nonnegative weight profile across directions.}
\label{fig:overview}
\end{figure}

AnisoFlow makes the prediction target learnable: starting from the standard
velocity target $\vu=\bm{\epsilon}-\vx$, it scales the noise component by a
separate nonnegative weight in each direction of a learned orthogonal basis,
while leaving the clean-data component intact.
A closed-form map recovers the full velocity from the network's prediction,
allowing AnisoFlow to retain the standard network architecture, flow-matching
loss, and sampler. We first construct the target and its recovery
(Fig.~\ref{fig:overview}), then show that standard targets arise as special
cases, and finally describe how the basis is learned.

\textbf{A learnable prediction target.}
We propose reweighting only the noise term of the velocity target and leaving its clean-data term intact. Scaling the data term as
well would be redundant, since an overall per-direction scale on the target is absorbed by the
network, so we fix the data coefficient to $-1$ and learn only the noise weights
(Appendix~\ref{app:theory}). We further ask the reweighting to admit exact velocity recovery at every
noise level $t\in(0,1]$, which holds precisely when the operator has no negative real eigenvalue. Combined with
symmetry, this restricts it to positive semidefinite (PSD) operators (Appendix~\ref{app:theory}). Concretely, we define the AnisoFlow target
\begin{equation}\label{eq:target}
\vu_W := \mW\bm{\epsilon} - \vx,
\end{equation}
where $\mW$ is a PSD operator. Writing the eigendecomposition
\begin{equation}
\mW = \mB\operatorname{diag}(w_1,\dots,w_D)\mB\T, \qquad w_i\ge0,
\end{equation}
with orthonormal basis $\mB=[\vb_1,\dots,\vb_D]$, where $D$ is the data dimension, the design space consists of the eigenvalues, which act as
per-direction noise weights, and the basis. In coordinates along $\mB$ the target reads
$u_{W,i}=w_i\epsilon_i - x_i$, so the data component remains full-dimensional while the noise
component receives a separate nonnegative weight in each direction. In patchified
models, $\mW$ acts on each patch (token) and is shared across spatial locations, so
$D$ is the patch dimension, e.g., $D=16\times16\times3=768$ for JiT
(Appendix~\ref{app:imagenet}).

\textbf{Velocity recovery.}
Given a prediction $\hat{\vu}_W$ of the AnisoFlow target, the velocity is recovered as
\begin{equation}\label{eq:recovery}
\hat{\vu} = (\alpha_t\mW + \sigma_t\mI)^{-1}\big((\mI-\mW)\vz_t + \hat{\vu}_W\big),
\qquad\text{equivalently}\qquad
\hat u_i = \frac{(1-w_i)z_{t,i} + \hat u_{W,i}}{\alpha_t w_i + \sigma_t}
\end{equation}
coordinate-wise in the eigenbasis of $\mW$ (Appendix~\ref{app:theory}). The denominator
$\alpha_t w_i + \sigma_t$ is positive on $t\in(0,1]$ whenever $w_i\ge0$, which is why we restrict to
nonnegative weights. This denominator also controls the amplification of
target-prediction error: for clean-data prediction ($w_i=0$), the amplification
scales as $1/\sigma_t$, making velocity recovery ill-conditioned at low noise,
whereas nonzero weights can improve its conditioning. The weights $w_i$ are shared across timesteps; in practice, allowing separate
weights across timestep buckets provides no measurable benefit
(Fig.~\ref{fig:bucket-weights}).

\textbf{Training objective.}
The network parameters $\theta$ and the weights $\vw=(w_1,\dots,w_D)$ are trained jointly with the recovered-velocity loss
$\mathcal L(\theta,\vw)=\E_{t,\vx,\bm{\epsilon}}\big[\|\hat{\vu}(\vz_t,t)-\vu\|^2\big]$,
where $\hat{\vu}$ is obtained from the network's prediction $\hat{\vu}_W$ through the recovery
map in Eq.~(\ref{eq:recovery}). The objective is the squared error on the \emph{recovered} velocity, not
directly on the AnisoFlow target
$\|\hat{\vu}_W-\vu_W\|^2$.

\textbf{Special cases.}
The family contains several standard targets as special cases. With $w_i=0$ the target in direction $i$ is clean-data
($x_i$) prediction up to sign; with $w_i=1$ it is the velocity $u_i=\epsilon_i-x_i$; in the PCA
basis of the data with $w_i\in\{0,1\}$ it reproduces the binary low-rank mask of AsymFlow
\citep{asymflow}; with $w_i\equiv w$ constant
across directions it becomes an isotropic scalar interpolation, equivalent up to scale and sign to
scalar target families such as k-Diff \citep{kdiff}; and the noise ($\epsilon_i$) prediction limit is obtained as
$w_i\to\infty$, again up to scale.

\textbf{Learnable basis.}
The basis $\mB$ is learned jointly with the network and the weights, and we consider two optimizers (Table~\ref{tab:target-ablation}). \emph{Stiefel Adam} treats $\mB$ as a parameter on the Stiefel manifold of orthonormal matrices and adapts Adam \citep{kingma2014adam} to it (Appendix~\ref{app:stiefel}). The \emph{spectral update} is a closed-form alternative: assuming the per-direction loss terms share a common eigenbasis, the optimal basis is given by the eigenvectors of a single moment matrix of the network's prediction errors, which we track with an exponential moving average and eigendecompose at every step (Proposition~\ref{prop:spectral}, Appendix~\ref{app:spectral}). In both cases the weights $w_i$ remain learned by gradient descent. We use the spectral update by default, as it achieves the best FID in our experiments.

\textbf{Initialization.}
By default we initialize $\mB=\mI$ and $w_i=1$, so that $\mW=\mI$ and the initial target is ordinary velocity prediction. Optionally, we use a \emph{PCA initialization}: $\mB$ is set to the PCA basis of the data and $w_i=c\lambda_i$, where $\lambda_i$ is the data variance along $\vb_i$ and $c=0.15$ matches the slope learned by our converged JiT-H/16 models. This choice is motivated by our analysis in \S\ref{sec:theory}, which singles out the PCA basis and predicts weights that scale near-linearly with $\lambda_i$. PCA initialization is not necessary: it speeds up early training empirically (Table~\ref{tab:target-ablation}), but the learned basis also converges close to the PCA basis on its own without PCA initialization (Fig.~\ref{fig:learned-w-b}). 

\section{Analysis of the Learned Weights}
\label{sec:theory}

We analyze how the learned weights relate to the data spectrum and why a single weight
profile suffices across timesteps. We study each direction of the PCA basis separately.
In this basis the data coordinates are uncorrelated, with variances
$\lambda_i=\E[x_i^2]$, while the Gaussian noise remains isotropic, so each weight $w_i$
is associated with a single direction. Full derivations are given in
Appendix~\ref{app:learned-weights}.

\textbf{Motivation for the PCA basis.}
Three observations motivate this choice. First, under a
linear-network model, in which the network is replaced by the optimal linear predictor,
the PCA basis is a fixed point of the spectral update (Section~\ref{sec:method}), whose
local stability we characterize in Appendix~\ref{app:linear-fixed-point}. Second, the
update recovers the PCA basis in expectation at initialization, when the network output
is zero (Appendix~\ref{app:spectral}). Third, the basis learned in practice closely
aligns with PCA (Fig.~\ref{fig:learned-w-b}).

\textbf{Error amplification and prediction difficulty.}
Because the velocity recovery map (Eq.~\ref{eq:recovery}) is invertible, for every $w_i$
the optimal prediction of the target $u_{W,i}$ recovers the same conditional-mean
velocity $\E[u_i\mid\vz_t,t]$. The
choice of $w_i$ therefore does not change the minimum achievable loss, only the function
the network must approximate to reach it. In direction $i$ at timestep $t$, the
recovered-velocity loss is
\begin{equation}\label{eq:recovered-loss}
\mathcal E_i(w_i;t)
=
\frac{\E\big[(\hat u_{W,i}-u_{W,i})^2\mid t\big]}
     {(\alpha_t w_i+\sigma_t)^2}.
\end{equation}
Increasing $w_i$ enlarges the denominator, so errors in the predicted target are
amplified less when converted back to velocity. A larger $w_i$, however, also
places more noise in the target, which the network must then predict. The learned
weight balances these two effects.

\textbf{Preferred weight at a fixed timestep.}
To make this trade-off concrete, we model the network's prediction as combining a noise
estimate $\hat\epsilon_i$ and a data estimate $\hat x_i$, so that
$\hat u_{W,i}\approx w_i\hat\epsilon_i-\hat x_i$, mirroring the target
$u_{W,i}=w_i\epsilon_i-x_i$. Even an optimal network cannot recover $\epsilon_i$ and $x_i$ exactly from $\vz_t$. The
best achievable estimates are the conditional means
$\epsilon_i^*=\E[\epsilon_i\mid\vz_t,t]$ and $x_i^*=\E[x_i\mid\vz_t,t]$, so we measure
the network's errors relative to them,
$A_i(t)=\E[(\hat\epsilon_i-\epsilon_i^*)^2\mid t]$ and
$C_i(t)=\E[(\hat x_i-x_i^*)^2\mid t]$.
The remaining uncertainty is unavoidable and does not depend on $w_i$. We treat
$A_i(t)$ and $C_i(t)$ as locally fixed, assuming that small changes in $w_i$ do not
change how accurately the network estimates the noise and the data, and neglect the
correlation between the two errors. If each timestep were optimized separately, the
preferred weight at timestep $t$ would then be
\begin{equation}\label{eq:preferred-weight}
w_i^*(t)=\frac{\alpha_t}{\sigma_t}\,\frac{C_i(t)}{A_i(t)}.
\end{equation}
Thus, directions where data estimation is relatively more difficult favor larger noise
weights, while directions where noise estimation is relatively more difficult favor
smaller weights.

\textbf{A single weight across timesteps.}
The explicit factor $\alpha_t/\sigma_t$ ranges from $0$ to $\infty$, suggesting that
different timesteps could prefer very different weights. The two estimation errors,
however, vary in the opposite direction. At small $t$, $\vz_t$ is dominated by the clean
data, making $x_i$ relatively easy to estimate while $\epsilon_i$ is difficult to infer.
At large $t$, this relation reverses. A simple sufficient condition for the two effects
to cancel is $A_i(t)=\alpha_t q_i(t)\bar A_i$ and $C_i(t)=\sigma_t q_i(t)\bar C_i$, where
$\bar A_i$ and $\bar C_i$ are direction-dependent scales and $q_i(t)\ge0$ is a shared
time dependence. Every timestep then prefers the same weight $w_i^*=\bar C_i/\bar A_i$,
which therefore also minimizes the loss averaged over timesteps that training optimizes.
Consistent with this analysis, weights learned independently across timestep buckets
nearly coincide, and allowing the weights to vary with time does not improve FID
(Fig.~\ref{fig:bucket-weights}).

\textbf{Near-linear spectral scaling.}
The shared weight $w_i^*=\bar C_i/\bar A_i$ is the ratio of the data- and
noise-estimation error scales in direction $i$. Two further assumptions connect it to
the data variance. Because the noise is isotropic, we take the noise-estimation error to
vary weakly across directions, $\bar A_i\approx\bar A$. We further assume the network
estimates each data coordinate with comparable relative error, so that the
data-estimation error scales with the data energy, $\bar C_i\propto\lambda_i$. Together,
these give $w_i^*=\bar C_i/\bar A_i\propto\lambda_i$, explaining the near-linear scaling
$w_i\approx c\lambda_i$ observed in pixel space (Fig.~\ref{fig:learned-w-a}).

\begin{table}[t]
\centering
\caption{\textbf{Effect of the learnable prediction target}, isolating noise weights/basis optimizers and data-dependent initialization. JiT-H/16 network.}
\label{tab:target-ablation}
\scalebox{0.9}{%
\begin{tabular}{lllccccc}
\toprule
\textbf{Method} & \textbf{Noise weights} & \textbf{Noise basis} & \textbf{Init} & \makecell{\textbf{FID}\textdownarrow \\ {\small @160ep}} & \makecell{\textbf{FID}\textdownarrow \\ {\small @600ep}} & \makecell{Training \\ sec/iter} & \makecell{Inference \\ sec/step} \\
\midrule
JiT       & Frozen (zeros) & -- & -- & 2.68 & 1.90 & 0.803 & 0.179 \\
AsymFlow  & Frozen (binary) & Frozen (PCA) & PCA & 2.34 & 1.76 & 0.803 & 0.179 \\
\midrule
AnisoFlow & Adam & Frozen (PCA) & PCA & \textbf{2.25} & 1.63 & 0.805 & 0.180 \\
AnisoFlow & Adam & Stiefel Adam & PCA & 2.32 & 1.68 & 0.818 & 0.180 \\
AnisoFlow & Adam & Spectral update & PCA & \underline{2.28} & 1.63 & 0.819 & 0.180 \\
AnisoFlow & Adam & Stiefel Adam & -- & \underline{2.28} & \underline{1.60} & 0.818 & 0.180 \\
AnisoFlow & Adam & Spectral update & -- & 2.30 & \textbf{1.55} & 0.819 & 0.180 \\
\bottomrule
\end{tabular}
}
\end{table}

\section{Experiments}
\label{sec:exp}

\textbf{Experimental setup.}
We evaluate AnisoFlow on class-conditional ImageNet~\citep{deng2009imagenet} $256{\times}256$ and text-to-image generation in pixel-space, and then further extend it to latent-space flow models.
In each setting, we preserve the architecture and training recipe of the
underlying model and change only the prediction-target parameterization.
For our main pixel-space experiments, we use JiT-H/16 \citep{jit}, with
JiT-B/16 for smaller-scale experiments. We report FID~\citep{heusel2017fid} under the ADM evaluation protocol
\citep{dhariwal2021diffusionbeatsgan}. Full implementation and
evaluation details are given in Appendix~\ref{app:impl}.

\subsection{Pixel-Space ImageNet Generation}

\subsubsection{Main results}

\begin{figure}[t]\centering
\captionsetup[subfigure]{skip=2pt}
\hfill
\begin{subfigure}[t]{0.353\linewidth}\centering
\includegraphics[width=0.9\linewidth]{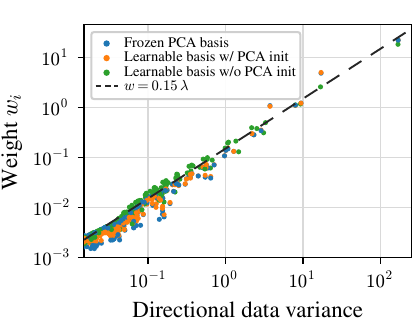}
\caption{Learned weights (trimmed range)}\label{fig:learned-w-a}
\end{subfigure}
\hfill
\begin{subfigure}[t]{0.529\linewidth}\centering
\includegraphics[width=0.9\linewidth]{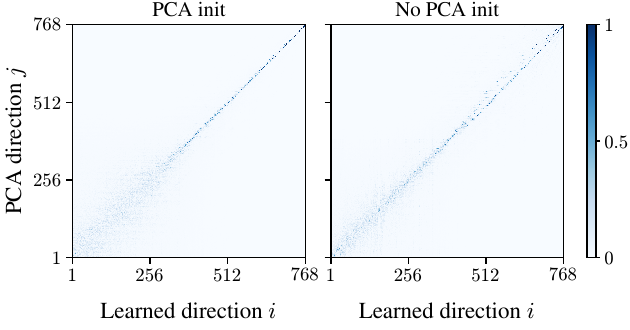}
\caption{Alignment between learned and PCA bases $\lvert \bm{v}_j\T \vb_i\rvert$}\label{fig:learned-w-b}
\end{subfigure}
\hfill
\caption{\textbf{Emergent spectral structure in AnisoFlow.}
\textbf{(a)}~Learned weights versus directional data variance
$\vb_i\T\mSigma\vb_i$, where $\mSigma$ is the data covariance, for a frozen PCA basis and learnable bases with
and without PCA initialization. The displayed range is trimmed to
variances $\geq0.015$.
\textbf{(b)}~Absolute inner products $\lvert \bm{v}_j\T \vb_i\rvert$ between PCA directions $\bm{v}_j$ and learned
directions $\vb_i$, in their native spectral order. Both initializations yield
near-diagonal alignment, strongest at high-variance
directions. All models are JiT-H/16; both learnable variants use the spectral
update in Table~\ref{tab:target-ablation}.}
\label{fig:learned-w}
\end{figure}

\begin{wraptable}{R}{0.5\linewidth}
\vspace{-1.5\baselineskip}
\centering
\caption{\textbf{ImageNet $256{\times}256$ pixel diffusion.} Best reported FID, grouped by architecture family. FLOP estimation follows \citet{yu2026pixeldit}, and * marks the JiT evaluation protocol, which can give up to $0.08$ lower FID than the ADM protocol. $\dagger$: uses a pretrained representation encoder (e.g., REPA-style alignment or DINO-based losses) or self-supervised pre-training.
}
\label{tab:imagenet}
\resizebox{\linewidth}{!}{%
\setlength{\tabcolsep}{0.3em}
\begin{tabular}{lcccl}
\toprule
\textbf{Method} & \textbf{Target} & \textbf{Params} & \textbf{GFLOPs} & \textbf{FID}\textdownarrow \\
\midrule
\multicolumn{5}{l}{\emph{Hierarchical CNNs (U-Net-like)}} \\
ADM-G & $\bm{\epsilon}$ & 554M & 2240 & 4.59 \\
\midrule
\multicolumn{5}{l}{\emph{Hierarchical transformers (U-ViT-like)}} \\
RIN               & $\bm{\epsilon}$ & 320M & 668  & 3.42 \\
SiD, UViT/2       & $\bm{\epsilon}$ & 2B   & 1110 & 2.44 \\
VDM++, UViT/2     & $\bm{\epsilon}$ & 2B   & 1110 & 2.12 \\
SiD2, UViT/2      & $\bm{\epsilon}$ & --   & 274  & 1.73 \\
PixelU-H/16$^\dagger$       & $\vx$           & 1.2B & 274  & 1.63* \\
EPG-G/16$^\dagger$          & $\vx$           & 1.4B & 642  & \underline{1.58} \\
SiD2, UViT/1      & $\bm{\epsilon}$ & --   & 1306 & \textbf{1.38} \\
\midrule
\multicolumn{5}{l}{\emph{Hierarchical transformers (DDT-like)}} \\
PixNerd-XL/16$^\dagger$     & $\bm{\epsilon}-\vx$ & 700M & 268 & 2.15 \\
DiP-XL/16$^\dagger$         & $\bm{\epsilon}-\vx$ & 631M & --  & 1.79 \\
DeCo-XL/16$^\dagger$        & $\bm{\epsilon}-\vx$ & 682M & 245 & 1.62 \\
PixelDiT-XL/16$^\dagger$    & $\bm{\epsilon}-\vx$ & 797M & 311 & 1.61 \\
FrequencyBooster-H/16$^\dagger$ & $\vx$ & 1.1B & -- & \underline{1.60}* \\
HyperDiT-H$^\dagger$        & $\bm{\epsilon}-\vx$ & 952M & 458 & \textbf{1.56}* \\
\midrule
\multicolumn{5}{l}{\emph{Plain transformers (DiT-like)}} \\
PixelFlow-XL/4    & $\bm{\epsilon}-\vx$ & 677M & 5818 & 1.98 \\
JiT-H/16          & $\vx$ & 953M & 364  & 1.86* \\
PixelGen-XL/16$^\dagger$    & $\vx$ & 676M & 260  & 1.83 \\
JiT-G/16          & $\vx$ & 2B   & 766  & 1.82* \\
PixelREPA-H/16$^\dagger$    & $\vx$ & 953M & 364  & 1.81* \\
AsymFlow-H/16$^\dagger$     & $\mP\bm{\epsilon}-\vx$ & 953M & 364 & \underline{1.57} \\
\textbf{AnisoFlow-H/16 (ours)} & $\mW\bm{\epsilon}-\vx$ & 953M & 365 & \textbf{1.55} \\
\bottomrule
\end{tabular}
}
\vspace{-1\baselineskip}
\end{wraptable}

\textbf{Learning the target, weights, and basis.}
Table~\ref{tab:target-ablation} isolates the effects of learning the weights, learning the basis, and initialization on a fixed JiT-H/16 backbone. For the basis, we compare the proposed Stiefel Adam and spectral update optimizers with a frozen PCA basis. Learning continuous
directional weights at a frozen PCA basis improves considerably over both clean-data prediction
(JiT, FID $1.90$) and AsymFlow's frozen binary mask (FID $1.76$), reaching
$1.63$ FID, showing that much of the gain comes from learning the directional
weights themselves. The frozen PCA basis also gives the best FID early in training ($2.25$). On the other hand, learning the basis without PCA initialization (isotropic $\mW=\mI$ initialization) eventually improves the final quality, with spectral update obtaining the best final FID of $1.55$. Despite different initialization and optimization paths, the learned basis remains close to PCA and the weights recover a PCA spectral profile, which we examine in \S\ref{sec:analysis}.

\textbf{Comparison with prior pixel diffusion models.}
In Table~\ref{tab:imagenet}, we compare prior pixel diffusion models on ImageNet $256{\times}256$ generation, grouped by architecture family: U-Net CNNs
\citep{dhariwal2021diffusionbeatsgan},
U-ViT-like hierarchical transformers
\citep{jabri2023rin,hoogeboom2023simplediffusion,kingma2023elbo,
hoogeboom2025simplerdiffusion,guo2026pixelu,lei2026novae}, DDT-like \citep{wang2026ddt} hierarchical transformers
\citep{wang2026pixnerd,chen2026dip,ma2026deco,yu2026pixeldit,
ma2026frequencybooster,he2026hyperdit},
and plain DiT-like transformers \citep{chen2025pixelflow,jit,ma2026pixelgen,shin2026repajit,asymflow}.
AnisoFlow-H/16 achieves the best in-class FID of $1.55$ among plain DiT-like pixel-space models, with only flow-matching training and no representation alignment (REPA)~\citep{repa} or extra pretraining. It also outperforms pixel models with complex hierarchical architectures and REPA (FID 1.56*).

\textbf{Compute overhead.}
Compared to JiT (Table~\ref{tab:target-ablation}), AnisoFlow adds only $0.6\%$ inference overhead and at most $2.0\%$ training overhead learning the weights and basis; Appendix~\ref{app:overhead} provides a
detailed breakdown.

\subsubsection{What does AnisoFlow learn?}
\label{sec:analysis}

\begin{figure}[t]
\centering

\begin{minipage}[t]{0.62\textwidth}
\vspace{0pt} %
\centering
\begin{minipage}[t]{0.42\linewidth}
\vspace{0pt}
\centering
\includegraphics[width=\linewidth]{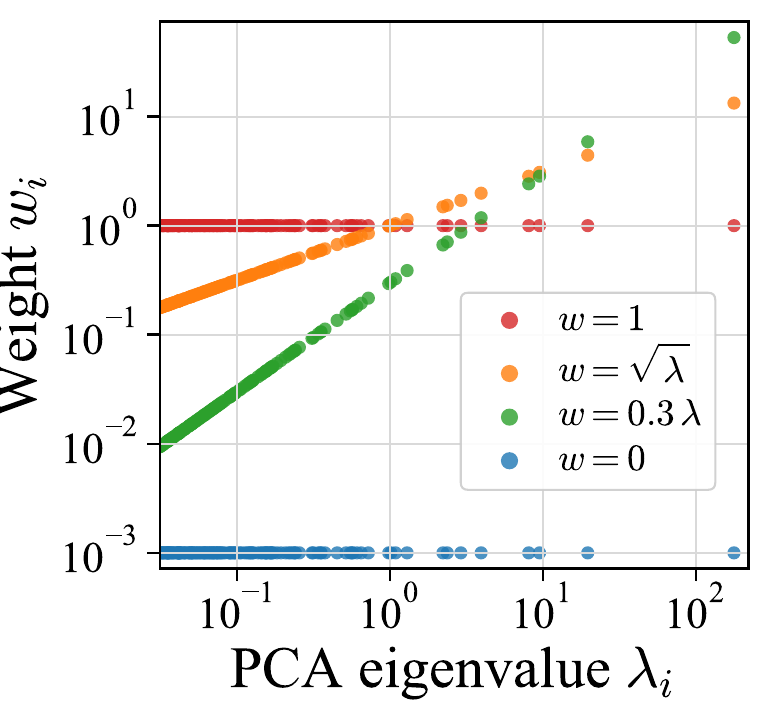}\\
{\footnotesize (a) Initialization}
\end{minipage}
\begin{minipage}[t]{0.42\linewidth}
\vspace{0pt}
\centering
\includegraphics[width=\linewidth]{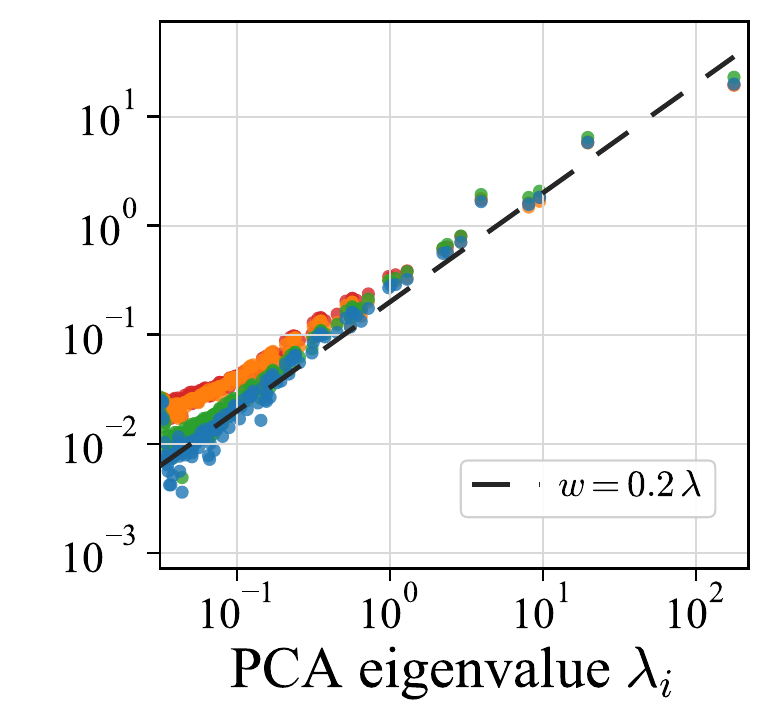}\\
{\footnotesize (b) Learned weights}
\end{minipage}

\caption{\textbf{Robustness to weight initialization.}
\textbf{(a)}~Four initial weight profiles and \textbf{(b)}~the learned weights after training
JiT-B/16 converge to nearly the same near-linear profile.}
\label{fig:init-robustness}
\end{minipage}
\hfill
\begin{minipage}[t]{0.31\textwidth}
\vspace{0pt} %
\centering
\includegraphics[width=\linewidth]{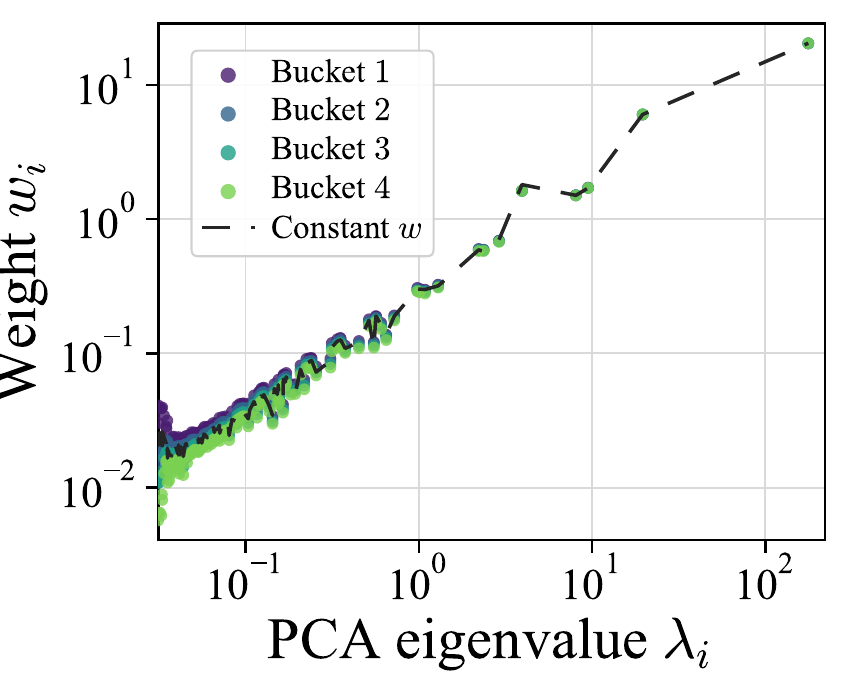}

\caption{\textbf{Time-bucketed weight profiles.} Four timestep-bucket profiles
(JiT-B/16) nearly coincide.}
\label{fig:bucket-weights}
\end{minipage}

\end{figure}

\begin{figure}[t]\centering
\includegraphics[width=\linewidth]{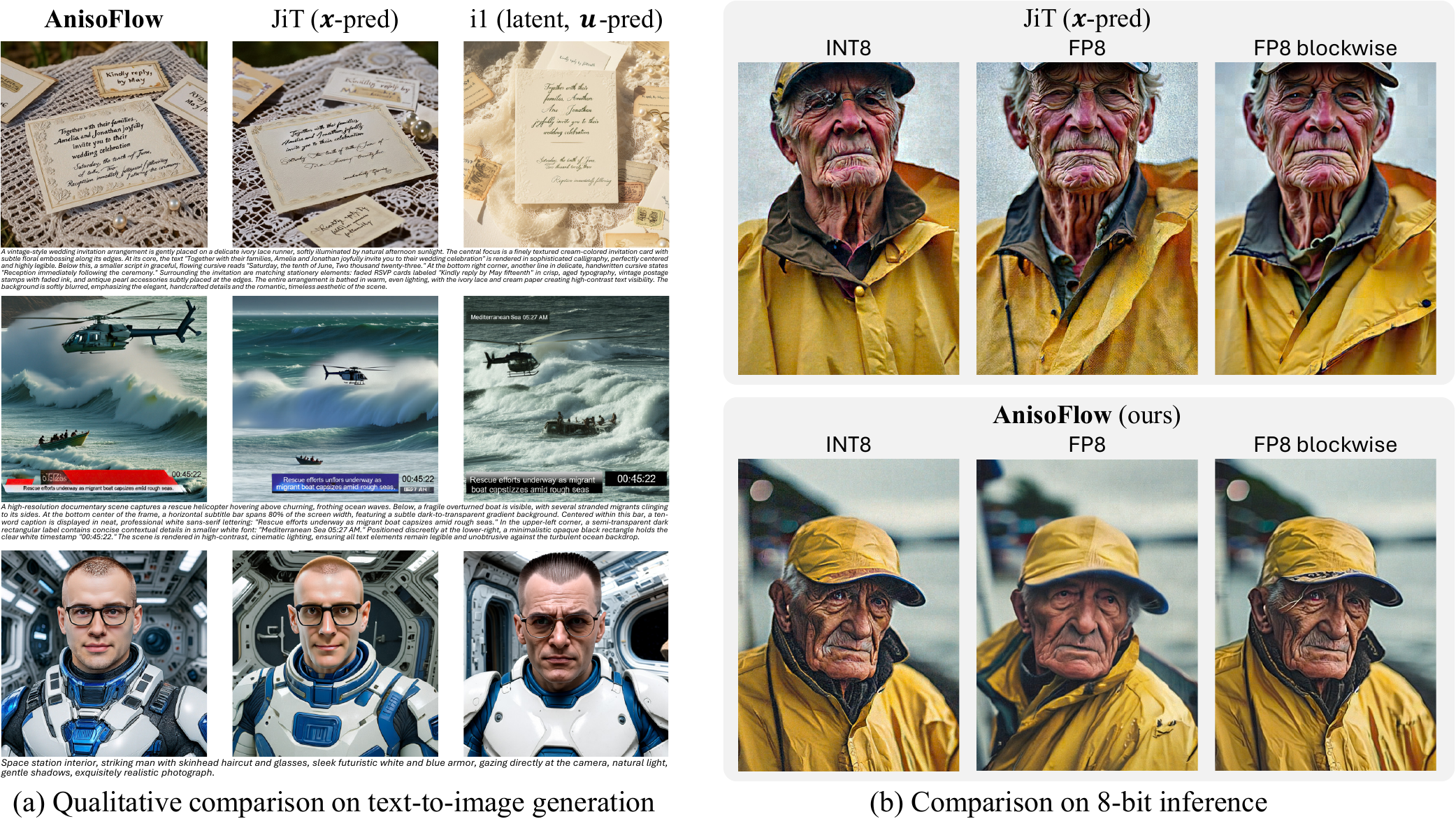}
\caption{\textbf{Qualitative text-to-image results.} \textbf{(a)}~$512{\times}512$ generations from AnisoFlow, the JiT-L/16-T2I baseline ($\vx$-prediction), and the latent i1-1B model ($\vu$-prediction), using identical prompts and sampling seeds. AnisoFlow renders finer texture than the $\vx$-prediction baseline (paper, fabric, waves, skin) and a sharper, more distinct style than the latent model, in line with its lower FID and pFID. \textbf{(b)}~The same prompt and seed under 8-bit inference (INT8, FP8, and blockwise FP8). $\vx$-prediction develops patch seams or strong noise, while AnisoFlow remains stable.}
\label{fig:t2i-comparison}
\end{figure}

\textbf{Learned weights follow the data spectrum.}
As shown in Fig.~\ref{fig:learned-w-a},
under different initialization and basis 
optimization paths, the learned directional weights of the pixel-space model always follow a near-linear relationship with the data variance along their corresponding
learned directions. Meanwhile, the basis learned from an isotropic initialization aligns closely with the PCA basis
(Fig.~\ref{fig:learned-w-b}). The directional variances therefore closely
track the PCA spectrum $\lambda_i$, recovering the near-linear scaling
$w_i\approx c\lambda_i$ predicted by our analysis.

\textbf{Effect of initialization.}
Initialization primarily affects the optimization trajectory rather than the
final spectral structure. On JiT-H/16, initializing at $\mW=\mI$ gives worse
FID early in training but the best FID after 600 epochs
(Table~\ref{tab:target-ablation}), while diverse initial weight profiles on
JiT-B/16 converge to similar near-linear solutions
(Fig.~\ref{fig:init-robustness}).

\textbf{A single weight profile suffices across timesteps.}
To test whether the weights need to vary with time, we split the timestep range
at the quartiles of the training timestep distribution and learn an independent
profile $w_i^{(b)}$ for each of the four buckets. The profiles remain close across
buckets and to the constant-$w_i$ solution
(Fig.~\ref{fig:bucket-weights}), while the added flexibility yields no FID
improvement ($4.92$ for shared $w_i$ vs.\ $4.96$ for time-dependent $w_i^{(b)}$ after 160 epochs). This supports using a single shared weight profile, consistent with
the cancellation analysis in \S\ref{sec:theory}.

\subsection{Pixel-Space Text-to-Image Generation}

We further apply AnisoFlow to pixel-space text-to-image generation using JiT-L/16-T2I as the baseline, which adopts the MM-JiT denoiser architecture from MiniT2I~\citep{minit2i}, trained with the text encoder, data, and training
schedule of i1~\citep{zeng2026i1}. Appendix~\ref{app:t2i} gives the training and evaluation details. We compare fixed $\vx$-prediction with the learned AnisoFlow target (with PCA initialization and spectral update) under matched training and evaluation. We evaluate at $512{\times}512$ on OneIG-Bench~\citep{chang2025oneig}, LongText-Bench~\citep{geng2025xomni}, GenEval~\citep{ghosh2023geneval}, DPG-Bench~\citep{hu2024ella}, HPSv3~\citep{ma2025hpsv3}, and COCO-10k~\citep{lin2014coco,asymflow} (FID and patch FID (pFID)~\citep{lin2024sdxllightning}).

\textbf{Main results.}
AnisoFlow improves over the $\vx$-prediction baseline on every metric except GenEval (Table~\ref{tab:t2i}). The largest gains are in HPSv3 ($7.71{\to}8.03$),
COCO-10k FID ($22.0{\to}20.1$), and pFID ($24.4{\to}22.7$), indicating a closer match to the style and fine detail of real images. AnisoFlow also surpasses the latent i1-1B model, which shares the text encoder and an equivalent training schedule but uses a larger denoiser, on the detail- and style-sensitive metrics LongText-Bench, HPSv3, FID, and pFID. Fig.~\ref{fig:t2i-comparison}(a) shows this qualitatively: AnisoFlow renders finer texture than the baseline and a sharper, more distinct style than the latent model, while achieving accurate text rendering.

\textbf{Reduced-precision stability.}
Reduced-precision inference is important for the efficiency of large generative models, but it makes numerical conditioning critical. For $\vx$-prediction, velocity recovery divides the prediction error by $\sigma_t$, amplifying quantization error at low noise levels into visible artifacts. AnisoFlow instead learns the prediction target that minimizes the recovered-velocity error directly, so it naturally converges to a well-conditioned target (\S\ref{sec:theory}). Fig.~\ref{fig:t2i-comparison}(b) compares INT8, FP8, and blockwise-FP8 \citep{deepseekai} inference: the baseline develops patch seams or strong noise, whereas AnisoFlow remains stable throughout.

\begin{table}[H]
\centering
\caption{\textbf{Text-to-image $512{\times}512$.} All models are trained using i1's data, text encoder, and schedule, and tested using the same guidance and sampler settings. * indicates using i1's rewritten prompts. Bold marks the best result within the matched pixel-space group.}
\label{tab:t2i}
{\footnotesize
\setlength{\tabcolsep}{4pt}
\scalebox{0.9}{%
\begin{tabular}{lcccccccc}
\toprule
\multirow{2}[2]{*}{\textbf{Method}} & \multirow{2}[2]{*}{\textbf{Params}} & \multirow{2}[2]{*}{\textbf{OneIG}\textuparrow} & \multirow{2}[2]{*}{\makecell[l]{\textbf{Long} \\ \textbf{Text}*}\textuparrow} & \multirow{2}[2]{*}{\makecell[l]{\textbf{Gen} \\ \textbf{Eval}*}\textuparrow} & \multirow{2}[2]{*}{\textbf{DPG}*\textuparrow} & \multirow{2}[2]{*}{\textbf{HPSv3}\textuparrow} & \multicolumn{2}{c}{COCO-10k} \\
\cmidrule(lr){8-9}
& & & & & & & \textbf{FID}\textdownarrow & \textbf{pFID}\textdownarrow \\
\midrule
i1-1B (FLUX.2 VAE, $\vu$-pred.) & 1.3B & 0.511 & 0.853 & 0.826 & 86.3 & 7.11 & 62.3 & 62.4 \\
\midrule
JiT-L/16-T2I ($\vx$-pred.) & 917M  & 0.512 & 0.862 & \textbf{0.820} & 85.7 & 7.71 & 22.0 & 24.4 \\
\textbf{AnisoFlow-L/16-T2I (ours)} & 918M & \textbf{0.516} & \textbf{0.864} & 0.814 & \textbf{85.8} & \textbf{8.03} & \textbf{20.1} & \textbf{22.7} \\
\bottomrule
\end{tabular}
}
}
\end{table}

\subsection{Latent-Space Models}
\label{sec:latent}

We also evaluate AnisoFlow on SiT-XL/2-REPA~\citep{ma2024sit,repa} and RAEv2~\citep{singh2026raev2}, preserving each model's
architecture and training/evaluation recipe while changing only the prediction
target, except that for RAEv2 both the baseline and AnisoFlow use a modified
timestep distribution and inference clamp that improve the baseline itself
(Appendix~\ref{app:impl}). AnisoFlow performs comparably to the default targets
(Table~\ref{tab:latent}), while the learned weights remain close to velocity
prediction for SiT-XL/2-REPA and to data prediction for RAEv2
(Fig.~\ref{fig:latent-weights}). The default targets thus appear well
matched to these representations. Unlike in pixel space, the latent weight
profiles are much flatter, so the assumptions behind the near-linear scaling do
not carry over directly.

\begin{table}[H]
\centering
\begin{minipage}[t]{0.54\linewidth}\vspace{0pt}\centering
\caption{\textbf{Latent-space results.}
AnisoFlow achieves comparable performance in latent-space models, where the
learned targets remain close to the respective default targets.
$^*$FID computed using JiT's reference statistics.}
\label{tab:latent}
\setlength{\tabcolsep}{3pt}
\renewcommand{\arraystretch}{1.30}
\scalebox{0.86}{%
\begin{tabular}{lccc}
\toprule
\textbf{Method} & \textbf{Target} & \textbf{FID}\textdownarrow & \textbf{JiT FID}*\textdownarrow \\
\midrule
SiT-XL/2-REPA                & velocity & 1.44 & 1.41 \\
\textbf{Aniso-SiT-XL/2-REPA (ours)} & learned & 1.44 & 1.40 \\
\midrule
RAEv2                     & $\vx$-pred. & 1.05 & 0.98 \\
\textbf{Aniso-RAEv2 (ours)} & learned & 1.05 & 0.96 \\
\bottomrule
\end{tabular}
}
\end{minipage}
\hfill
\begin{minipage}[t]{0.42\linewidth}\vspace{0pt}\centering
\captionsetup{type=figure,skip=2pt,justification=raggedright,singlelinecheck=false}
\makebox[\linewidth][l]{\hspace*{0.01\linewidth}%
\includegraphics[width=0.81\linewidth]{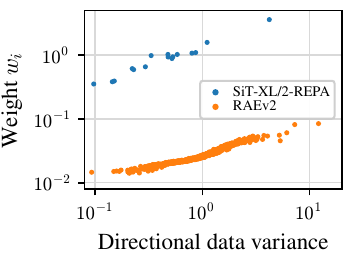}}
\caption{\textbf{Latent-space weights.} Learned weights in SiT-REPA and RAEv2 are close to their default targets.}
\label{fig:latent-weights}
\end{minipage}
\end{table}

\section{Conclusion}

AnisoFlow replaces fixed prediction targets with a learnable anisotropic target,
assigning a separate noise weight to each direction in a learned basis. In pixel
space, the learned basis aligns with PCA and the weights scale
near-linearly with data variance, consistent with our prediction-error analysis.
Empirically, AnisoFlow substantially improves FID over the JiT baseline on
ImageNet without representation alignment or architectural changes, while
improving image quality and numerical stability in text-to-image generation. These results show that prediction targets can adapt
to the data and the model's errors rather than being fixed in advance.

\textbf{Limitations.}
Our analysis uses a local model of prediction error and does not capture
full training dynamics or determine the constant in the observed spectral
scaling. Empirical gains are largest in pixel space and smaller in latent
space, where learned targets remain close to existing defaults.

\ificlrfinal
\textbf{Acknowledgments.}
This work was done while Rayhan Zirvi was a visiting undergraduate researcher at Stanford University, supported by the Summer Undergraduate Research Fellowships (SURF) program at Caltech.
\fi

\section*{AI Use Statement}
In this work, we used generative AI tools (LLM-based assistants) to edit the paper for grammar and readability, to help verify and refine the derivations in Section~5 and Appendix~A, and to accelerate the implementation of our experiments\ifarxiv\else{} and the supplementary code\fi. All AI-assisted work was reviewed and verified by the authors, and we take responsibility for the final content of this paper.

\section*{Reproducibility Statement}
\ifarxiv\else
We provide a minimal, self-contained PyTorch implementation of AnisoFlow for
ImageNet $256{\times}256$ in the supplementary material. It includes the
AnisoFlow target and velocity recovery, the spectral basis update, training and
sampling scripts, and the configuration of our main JiT-H/16 model. We will
release the full code and trained models upon publication.
\fi
Section~\ref{sec:method} and Appendix~\ref{app:param} describe how the
directional weights and basis are parameterized and optimized.
Appendices~\ref{app:imagenet}--\ref{app:latent-exp} give the training and
evaluation settings for the ImageNet, text-to-image, and latent-space
experiments, including the selected guidance settings
(Table~\ref{tab:imagenet-cfg}) and the text-to-image training stages
(Table~\ref{tab:t2i-hparams}), and Appendix~\ref{app:overhead} details the
compute overhead. For the analysis, Appendix~\ref{app:theory} states the
assumptions and gives full derivations, including the proof of
Proposition~\ref{prop:spectral}.

\bibliographystyle{iclr2027_conference}
\bibliography{references}

\clearpage
\appendix

\section{Theoretical Details}
\label{app:theory}

\subsection{Target Family and Recovery}
\label{app:target-family}

\textbf{Normalizing the linear target family.}
We first show that a general linear target has only one meaningful relative degree of freedom per
direction, since any nonzero overall scaling can be absorbed into the network output. In a single
direction, such a target combines the clean-data and noise coordinates,
$y=\varphi x+\psi\epsilon$. Rescaling the coefficients
$(\varphi,\psi)\mapsto(c\varphi,c\psi)$ with $c\neq0$ rescales the target by $c$. Any prediction
$f(z_t,t)$ for the original target can therefore be converted into a prediction $cf(z_t,t)$ for the
rescaled target by multiplying the network output by $c$, so this rescaling does not change the set
of functions the model can represent. Under squared error, the optimal prediction transforms in the
same way, $\E[c\,y\mid z_t,t]=c\,\E[y\mid z_t,t]$. We therefore fix the
overall scale by setting $\varphi=-1$, leaving the single relative noise weight $w=-\psi/\varphi$
and the AnisoFlow parameterization $u_{W,i}=w_i\epsilon_i-x_i$. The remaining case $\varphi=0$, pure
noise prediction, is recovered as $w_i\to\infty$ up to the overall scale already removed. Scaling
the data component in addition to the noise component thus introduces no new target family, though
different scalings may still affect gradients, optimization, or regularization in practice.

\textbf{Extension to mixed directions.}
The same argument applies to linear targets that mix directions. Given a predictor
$f(\vz_t,t)$ for a target $\vy$, composing its output with an invertible linear transformation
$\mT$ gives a predictor for the transformed target $\mT\vy$, and the optimal predictor transforms
in the same way, $\E[\mT\vy\mid\vz_t,t]=\mT\,\E[\vy\mid\vz_t,t]$. Applying $\mT=-\mK^{-1}$ to any
linear target $\vy=\mK\vx+\mL\bm{\epsilon}$ with invertible $\mK$ gives the normal form
$\vu_W=\mW\bm{\epsilon}-\vx$ with $\mW=-\mK^{-1}\mL$. We use $\mW$ to parameterize the linear
transformation applied to the noise relative to the fixed data coefficient $-\mI$. Targets with
singular $\mK$, such as pure noise prediction, fall outside this normal form, as in the scalar
case $\varphi=0$. As in the scalar case, the normalization preserves the set of representable
predictors up to an invertible output transformation, but it may change the geometry of the
training loss and hence optimization. The resulting $\mW$ need not be symmetric; we next derive the
recovery for general $\mW$ and then characterize which choices are admissible.

\textbf{Velocity recovery.}
We now assume the linear rectified-flow schedule $\alpha_t=1-t$ and $\sigma_t=t$, so that
$\alpha_t+\sigma_t=1$ and the flow-matching velocity is $u=\epsilon-x$. We first derive the
recovery formula in a single direction with scalar weight $w$, and then extend it to matrix form.
The interpolation and target are
\begin{equation}
z_t=\alpha_t x+\sigma_t\epsilon,
\qquad
u_W=w\epsilon-x .
\end{equation}
Using $\alpha_t+\sigma_t=1$ and collecting the data and noise coefficients,
\begin{equation}
(1-w)z_t+u_W
=-\big(\alpha_t w+\sigma_t\big)x+\big(\alpha_t w+\sigma_t\big)\epsilon
=\big(\alpha_t w+\sigma_t\big)(\epsilon-x).
\end{equation}
Since $u=\epsilon-x$, this gives $u=\big((1-w)z_t+u_W\big)/(\alpha_t w+\sigma_t)$. Replacing the
exact target by the network prediction yields the per-direction recovery formula of
Eq.~(\ref{eq:recovery}).
In matrix form, using
$\vz_t=\alpha_t\vx+\sigma_t\bm{\epsilon}$ and $\vu_W=\mW\bm{\epsilon}-\vx$, the same calculation gives
\begin{equation}
(\mI-\mW)\vz_t+\vu_W=(\alpha_t\mW+\sigma_t\mI)(\bm{\epsilon}-\vx)=(\alpha_t\mW+\sigma_t\mI)\,\vu ,
\end{equation}
so for a predicted target $\hat{\vu}_W$ inverting $\alpha_t\mW+\sigma_t\mI$ yields the matrix
recovery of Eq.~(\ref{eq:recovery}), valid whenever this operator is invertible, with no symmetry
required.
Applying this linear map before computing the loss lets us retain the usual
velocity-space flow-matching objective and sampler, without introducing additional network modules.
The recovery formula also determines which choices of $\mW$ keep the recovery nonsingular across
all timesteps.

\textbf{Why we restrict $\mW$ to be positive semidefinite.}
Recovery fails at time $t$ exactly when $-t/(1-t)$ is an eigenvalue of $\mW$, and this quantity
sweeps the entire negative real axis as $t$ ranges over $(0,1)$. Recovery is therefore nonsingular
at every interior timestep exactly when $\mW$ has no eigenvalue on the negative real axis; for
example, $\mW=-\mI$ becomes singular at $t=1/2$. This fully general family is algebraically possible
but not a practical target family. A general real operator may have complex eigenvalues and, when
non-normal, may be poorly conditioned even if its eigenvalues avoid the negative real
axis. It also does not admit the simple orthogonal, direction-wise recovery used by AnisoFlow. Restricting to symmetric operators makes all eigenvalues real, and the nonsingularity
condition then reduces to requiring nonnegative eigenvalues, i.e., $\mW\succeq 0$. A general PSD operator is
$\mW=\mB\operatorname{diag}(w_i)\mB\T$ with $w_i\ge0$ for an orthonormal basis $\mB$, the
parameterization of \S\ref{sec:method}, so we need only choose nonnegative eigenvalues and an
orthonormal basis. In this basis the recovery acts direction by direction, with the per-direction denominators
$\alpha_t w_i+\sigma_t$ of the scalar form above. These denominators are positive on $t\in(0,1]$,
and a zero weight is singular only at the data endpoint $t=0$, which we treat as a limiting value.
Our samplers never evaluate $t=0$, and the numerical safeguards on the recovery denominators are
given in Appendix~\ref{app:impl}.

\subsection{Analysis of the Learned Weights}
\label{app:learned-weights}
Throughout this subsection the basis is fixed to the PCA directions, which
closely match the learned basis (Fig.~\ref{fig:learned-w-b}).
Appendix~\ref{app:linear-fixed-point} studies when this alignment is preserved
and locally stable under the spectral update. Here we study the weights $w_i$. As in the main text, $w_i$
denotes the single weight used by AnisoFlow in direction $i$, shared across
timesteps, while $w_i^*(t)$ denotes the weight preferred at a
fixed timestep $t$ if that timestep were optimized in isolation.

\textbf{How prediction error enters the loss.}
Let $u_i^*(\vz_t,t)=\E[u_i\mid\vz_t,t]$ denote the conditional-mean velocity. For
any $w_i$, invertibility of the recovery map gives the corresponding optimal
target predictor
\begin{equation}
u_{W,i}^*(w_i;\vz_t,t)
=
(\alpha_t w_i+\sigma_t)u_i^*(\vz_t,t)
-(1-w_i)z_{t,i},
\end{equation}
which is exactly $\E[u_{W,i}\mid\vz_t,t]$. After recovery, this predictor yields
$u_i^*$, so the minimum achievable velocity-prediction loss is the same for
every $w_i$.
Let $e_{W,i}(w_i)=\hat u_{W,i}-u_{W,i}$ denote the network's error relative to the target
$u_{W,i}=w_i\epsilon_i-x_i$. The corresponding recovered-velocity error is
$e_{W,i}(w_i)/(\alpha_t w_i+\sigma_t)$,
so the recovered-velocity loss in direction $i$ is exactly Eq.~(\ref{eq:recovered-loss}).
Increasing $w_i$ enlarges the recovery denominator and therefore reduces
the amplification of target-prediction error, but it also changes the function the network must
approximate. The learned weight balances these two effects.
The loss in Eq.~(\ref{eq:recovered-loss}) decomposes as
\begin{equation}\label{eq:loss-decomposition}
\mathcal E_i(w_i;t)
=
\underbrace{
\frac{\E[(\hat u_{W,i}-u_{W,i}^*)^2\mid t]}
     {(\alpha_t w_i+\sigma_t)^2}
}_{\text{excess prediction error}}
+
\underbrace{
\E[(u_i-u_i^*)^2\mid t]
}_{\text{irreducible error}}.
\end{equation}
By orthogonality of the conditional mean, the second term is the minimum
achievable velocity-prediction loss and is independent of $w_i$. Thus, only the
excess prediction error can affect which weight is preferred.

\textbf{A local frozen-error model.}
To understand how the loss varies with $w_i$ near a trained solution, we use a local
frozen-error model. Let $\epsilon_i^*=\E[\epsilon_i\mid\vz_t,t]$ and
$x_i^*=\E[x_i\mid\vz_t,t]$, so that the optimal target predictor is
$u_{W,i}^*=w_i\epsilon_i^*-x_i^*$. Around the current learned weight, we model the
network's prediction as if it were formed from a noise estimate $\hat\epsilon_i$ and a
data estimate $\hat x_i$, according to $\hat u_{W,i}(w_i)\approx w_i\hat\epsilon_i-\hat x_i$,
and measure their errors relative to the conditional means,
$e_{n,i}=\hat\epsilon_i-\epsilon_i^*$ and $e_{x,i}=\hat x_i-x_i^*$. The excess target error
is then $\hat u_{W,i}-u_{W,i}^*\approx w_i e_{n,i}-e_{x,i}$. Measuring errors relative to
the conditional means separates out the irreducible uncertainty, whose data and noise
components are perfectly anti-correlated given $z_{t,i}$, since
$\alpha_t(x_i-x_i^*)+\sigma_t(\epsilon_i-\epsilon_i^*)=0$, and which contributes only the
$w_i$-independent term in Eq.~(\ref{eq:loss-decomposition}).
We assume that, over the range of $w_i$ relevant near convergence, the conditional second-order
statistics of $e_{n,i}$ and $e_{x,i}$ vary weakly with $w_i$, and therefore treat them as fixed when analyzing
the dependence of the loss on $w_i$. Under this frozen-error approximation,
\begin{equation}
\E\big[(\hat u_{W,i}-u_{W,i}^*)^2\mid t\big]=A_i(t)w_i^2-2\Gamma_i(t)w_i+C_i(t),
\end{equation}
with
\begin{equation}
A_i(t)=\E[e_{n,i}^2\mid t],\qquad
\Gamma_i(t)=\E[e_{n,i}e_{x,i}\mid t],\qquad
C_i(t)=\E[e_{x,i}^2\mid t],
\end{equation}
and the fixed-timestep recovered-velocity loss becomes
\begin{equation}\label{eq:recovered-loss-quad}
\mathcal E_i(w_i;t)
=
\frac{A_i(t)w_i^2-2\Gamma_i(t)w_i+C_i(t)}
     {(\alpha_t w_i+\sigma_t)^2}
+\E[(u_i-u_i^*)^2\mid t],
\end{equation}
where the last term is independent of $w_i$. The corresponding time-averaged loss is
\begin{equation}\label{eq:time-averaged-loss}
\overline{\mathcal E}_i(w_i)=\E_t[\mathcal E_i(w_i;t)].
\end{equation}

\textbf{Fixed-timestep optima.}
For a fixed timestep, differentiating the first term of
Eq.~(\ref{eq:recovered-loss-quad}) gives
\begin{equation}
\frac{\partial}{\partial w}
\frac{A_i(t)w^2-2\Gamma_i(t)w+C_i(t)}
     {(\alpha_t w+\sigma_t)^2}
=
\frac{
2\left[
w\big(A_i(t)\sigma_t+\alpha_t \Gamma_i(t)\big)
-
\big(\Gamma_i(t)\sigma_t+\alpha_t C_i(t)\big)
\right]
}{
(\alpha_t w+\sigma_t)^3
}.
\end{equation}
When $A_i(t)\sigma_t+\alpha_t\Gamma_i(t)\neq0$, an interior stationary point satisfies
\begin{equation}
w_i^*(t)
=
\frac{
\Gamma_i(t)\sigma_t+\alpha_t C_i(t)
}{
A_i(t)\sigma_t+\alpha_t \Gamma_i(t)
}.
\end{equation}
It is an interior minimum over $w\ge0$ when $A_i(t)\sigma_t+\alpha_t\Gamma_i(t)>0$ and $\Gamma_i(t)\sigma_t+\alpha_t C_i(t)>0$. Otherwise the stationary point need not minimize the loss on $w\ge0$.
When the cross terms are negligible, specifically
$|\alpha_t\Gamma_i(t)|\ll A_i(t)\sigma_t$ and
$|\sigma_t\Gamma_i(t)|\ll\alpha_t C_i(t)$, the preferred weight reduces to
\begin{equation}
w_i^*(t)
\approx
\frac{\alpha_t}{\sigma_t}
\frac{C_i(t)}{A_i(t)}.
\end{equation}
Under $\Gamma_i(t)=0$, $A_i(t)>0$, $C_i(t)>0$, and $t\in(0,1)$, this stationary
point is the unique minimizer of the timestep loss over $w\ge0$.

\textbf{When a constant weight suffices.}\label{app:constant-w}
Under $\Gamma_i(t)=0$, the timestep-wise preferred weight
$w_i^*(t)=(\alpha_t/\sigma_t)\,C_i(t)/A_i(t)$
is nominally time dependent through both the schedule ratio
$\alpha_t/\sigma_t$ and the error ratio $C_i(t)/A_i(t)$. These two dependencies, however, naturally vary in opposite directions. At small
$t$, $\sigma_t$ is small and $\vz_t$ is dominated by the clean data, making $x_i$ relatively easy to
predict but $\epsilon_i$ difficult to infer. At large $t$, $\vz_t$ is increasingly dominated by
$\bm{\epsilon}$, so estimating $\epsilon_i$ becomes easier while estimating $x_i$ becomes harder.
Accordingly, the data-estimation error $C_i(t)$ tends to increase with $t$, while the
noise-estimation error $A_i(t)$ tends to decrease.
This suggests that the time dependence in $A_i(t)$ and $C_i(t)$ may cancel the explicit schedule
factor in $w_i^*(t)$. Consistent with this picture, in our time-dependent ablation, independently
learned weights across several timestep buckets converge to very similar values, and
allowing the weights to vary across timesteps does not improve FID (Fig.~\ref{fig:bucket-weights}). We therefore
consider the separable form
\begin{equation}\label{eq:separable}
A_i(t)=\alpha_t q_i(t)\bar A_i,
\qquad
C_i(t)=\sigma_t q_i(t)\bar C_i.
\end{equation}
The key assumption is that, after the factors $\alpha_t$ and
$\sigma_t$, the two error powers share the same remaining time dependence
$q_i(t)\ge0$, with $\bar A_i,\bar C_i>0$.

\textbf{Shared minimizer under the separable form.}
With $\Gamma_i(t)=0$, the per-timestep derivative becomes
\begin{align}
\frac{\partial}{\partial w}\frac{A_i(t)w^2+C_i(t)}{(\alpha_t w+\sigma_t)^2}
&=\frac{2\left[wA_i(t)\sigma_t-\alpha_t C_i(t)\right]}{(\alpha_t w+\sigma_t)^3} \\
&=\frac{2\alpha_t\sigma_t q_i(t)\bar A_i}{(\alpha_t w+\sigma_t)^3}\left(w-\frac{\bar C_i}{\bar A_i}\right).
\end{align}
Thus every interior timestep with $q_i(t)>0$ is minimized at the same weight
\begin{equation}\label{eq:shared-minimizer}
w_i^*=\frac{\bar C_i}{\bar A_i},
\end{equation}
while timesteps with $q_i(t)=0$ are indifferent to the weight. Therefore, when AnisoFlow uses a single weight shared across timesteps,
choosing this common value $w_i^*=\bar C_i/\bar A_i$ also minimizes the
time-averaged loss $\overline{\mathcal E}_i(w_i)$ under any timestep
distribution supported on interior timesteps, provided the expected loss is
finite. The minimizer is unique if $q_i(t)>0$ on a set of positive probability.
A time-dependent parameterization $w_i(t)$ is therefore unnecessary under
these conditions.

\textbf{Near-linear scaling across directions.}
To characterize how the learned optimum varies across directions, we make two additional
assumptions. Because the sampled noise is isotropic, we assume that the
direction-dependent factor $\bar A_i$ in the noise-estimation error is roughly constant across
directions, $\bar A_i\approx\bar A$. We further assume that the
direction-dependent factor $\bar C_i$ in the data-estimation error scales
roughly with the data variance. Since the data energy in direction
$i$ is $\E[x_i^2]=\lambda_i$, this gives $\bar C_i\propto\lambda_i$. The resulting optimum therefore
satisfies
\begin{equation}
w_i^*=\frac{\bar C_i}{\bar A_i}\propto\lambda_i.
\end{equation}
Our analysis does not predict the constant $c$ in the observed near-linear scaling
$w_i\approx c\lambda_i$, which depends on the network's error structure, the timestep sampler, and
optimization. Departures from these assumptions can change the prefactor and lead to deviations from
exact linear scaling.

\textbf{The zero-skill limit.}
We consider a contrasting limiting case in which the network has no usable information
about the realization of direction $i$ beyond the timestep $t$, so that its target
prediction is $\hat u_{W,i}=0$ (the recovery map still uses $z_{t,i}$, so only the learned
prediction is uninformative). For centered data independent of the noise, the
recovered-velocity loss is then
\begin{equation}\label{eq:zero-skill-loss}
\overline{\mathcal E}_i(w_i)
=
\E_t\left[
\frac{\E[(w_i\epsilon_i-x_i)^2]}{(\alpha_t w_i+\sigma_t)^2}
\right]
=
\E_t\left[
\frac{w_i^2+\lambda_i}{(\alpha_t w_i+\sigma_t)^2}
\right].
\end{equation}
Unlike the separable structure of Eq.~(\ref{eq:separable}), the numerator does not vary
with $t$, so the timestep-cancellation argument above does not apply, and the preferred
constant weight can depend on the timestep distribution. If the weight were
instead allowed to vary independently with time, minimizing the integrand at
each timestep gives
\begin{equation}\label{eq:zero-skill-pointwise}
w_i^*(t)=\frac{\alpha_t}{\sigma_t}\lambda_i.
\end{equation}
Thus, without structure in the network's prediction errors, the preferred
target retains an explicit timestep dependence.

\textbf{Connection to EDM preconditioning.}
The pointwise zero-skill solution coincides with a direction-wise extension of
EDM output preconditioning~\citep{edm}. At a fixed timestep $t<1$, so that
$\alpha_t>0$, divide the flow state by $\alpha_t$ and write, in the PCA basis,
\begin{equation}
\tilde z_{t,i}=x_i+s_t\epsilon_i,
\qquad
s_t=\frac{\sigma_t}{\alpha_t}.
\end{equation}
EDM parameterizes the clean-data estimate as $c_{\mathrm{skip}}\tilde z_t+c_{\mathrm{out}}F$,
where $F$ is the raw network output, and chooses $c_{\mathrm{skip}}$ to minimize
the variance of the residual $x-c_{\mathrm{skip}}\tilde z_t$ left for the network. This
makes $c_{\mathrm{skip}}\tilde z_t$ the linear least-squares estimate of $x$ from $\tilde z_t$. Replacing EDM's scalar data variance by the variance
$\lambda_i$ of direction $i$ gives
\begin{equation}
c_{\mathrm{skip},i}
=
\frac{\lambda_i}{\lambda_i+s_t^2}.
\end{equation}
The residual can then be written as
\begin{align}
x_i-c_{\mathrm{skip},i}\tilde z_{t,i}
&=
(1-c_{\mathrm{skip},i})
\left(x_i-w_i\epsilon_i\right),\\
w_i
&=
\frac{c_{\mathrm{skip},i}s_t}
     {1-c_{\mathrm{skip},i}}
=
\frac{\lambda_i}{s_t}
=
\frac{\alpha_t}{\sigma_t}\lambda_i.
\end{align}
Up to the timestep-dependent factor $-(1-c_{\mathrm{skip},i})$, this residual is the
AnisoFlow target $u_{W,i}=w_i\epsilon_i-x_i$; EDM's additional output scaling
$c_{\mathrm{out}}$ changes only this overall scale and not the relative
weight $w_i$ (Appendix~\ref{app:target-family}), and its input scaling does not
affect the target. Hence a direction-wise
extension of EDM yields exactly the pointwise zero-skill optimum in
Eq.~(\ref{eq:zero-skill-pointwise}). This correspondence is intuitive: EDM's
derivation uses only the second-order statistics of the data and noise, which
is precisely the information available to a network with no predictive skill.
Our general analysis instead depends on the network's learned error statistics
$A_i(t)$, $C_i(t)$, and $\Gamma_i(t)$, whose timestep dependence can offset
the explicit $\alpha_t/\sigma_t$ factor and favor a single weight profile
shared across timesteps (Eq.~(\ref{eq:separable}); Fig.~\ref{fig:bucket-weights}).

\subsection{Empirical Analysis of Prediction Errors}
\label{app:error-validation}

We empirically investigate the prediction-error structure underlying the preferred-weight analysis of Appendix~\ref{app:learned-weights}, using a trained AnisoFlow model on ImageNet.

\textbf{Experimental setup.}
We use the 600-epoch AnisoFlow-H/16 checkpoint with a frozen PCA basis. Since AnisoFlow predicts the combined target $w_i\epsilon_i-x_i$ through a single output head, its prediction cannot be uniquely decomposed into separate noise and data estimates. Our analysis therefore provides a consistency check of the theoretical explanation rather than a direct measurement of the underlying error components. We freeze the network and fit two auxiliary linear readouts to its intermediate features, producing estimates $\hat\epsilon_i$ and $\hat x_i$. Since the original output head is itself a linear projection of these features, linear readouts provide a natural way to estimate data and noise separately without modifying the learned representation. The readouts are fitted using weighted ridge regression on 30K ImageNet training images and evaluated on a disjoint set of 30K images.

We select 60 PCA directions across the 768-dimensional patch spectrum and restrict the analysis to 18 directions for which the auxiliary readouts accurately reconstruct the original model's predictions (relative reconstruction discrepancy below 5\% and recovered-velocity MSE increase below 1\%). These criteria assess reconstruction fidelity rather than agreement with the predicted scaling. The retained directions span nearly three orders of magnitude in directional variance, from $\lambda_i=0.205$ to $171$.

\textbf{Measuring the error statistics.}
The theoretical quantities $A_i(t)$, $C_i(t)$, and $\Gamma_i(t)$ in Appendix~\ref{app:learned-weights} are defined relative to the conditional means of the data and noise, which are not directly available for ImageNet. Instead, we measure residuals against the sampled data and noise:
\begin{equation}
r_{n,i}=\hat\epsilon_i-\epsilon_i,\qquad
r_{x,i}=\hat x_i-x_i.
\end{equation}
For each direction and timestep, we estimate the second moments
\begin{equation}
\begin{aligned}
a_i(t)&=\E[r_{n,i}^2\mid t],\\
b_i(t)&=\E[r_{x,i}^2\mid t],\\
g_i(t)&=\E[r_{n,i}r_{x,i}\mid t],
\end{aligned}
\end{equation}
averaging over evaluation images and spatial patches.

Let $x_i^*=\E[x_i\mid\vz_t,t]$ and define the irreducible posterior uncertainty $F_i(t)=\E[(x_i-x_i^*)^2\mid t]$, with class conditioning implicit. Since $\vz_t=\alpha_t\vx+\sigma_t\bm{\epsilon}$, the posterior uncertainties satisfy
\begin{equation}
\epsilon_i-\epsilon_i^*
=-\frac{\alpha_t}{\sigma_t}(x_i-x_i^*).
\end{equation}
Using the orthogonality of conditional-mean residuals, the observed moments decompose as
\begin{equation}
\begin{aligned}
a_i(t)&=A_i(t)+\frac{\alpha_t^2}{\sigma_t^2}F_i(t),\\
b_i(t)&=C_i(t)+F_i(t),\\
g_i(t)&=\Gamma_i(t)-\frac{\alpha_t}{\sigma_t}F_i(t).
\end{aligned}
\end{equation}

We can therefore construct two effective error statistics:
\begin{equation}
\begin{aligned}
A_i^{\mathrm{eff}}(t)
&=a_i(t)+\frac{\alpha_t}{\sigma_t}g_i(t)
=A_i(t)+\frac{\alpha_t}{\sigma_t}\Gamma_i(t),\\
C_i^{\mathrm{eff}}(t)
&=b_i(t)+\frac{\sigma_t}{\alpha_t}g_i(t)
=C_i(t)+\frac{\sigma_t}{\alpha_t}\Gamma_i(t).
\end{aligned}
\label{eq:effective-errors}
\end{equation}
The irreducible posterior uncertainty cancels exactly. Furthermore, the full preferred-weight expression from Appendix~\ref{app:learned-weights} becomes
\begin{equation}
w_i^*(t)
=\frac{\alpha_t}{\sigma_t}
\frac{C_i^{\mathrm{eff}}(t)}{A_i^{\mathrm{eff}}(t)},
\label{eq:effective-preferred-weight}
\end{equation}
for an interior optimum under the local frozen-error model. This expression accounts for the cross moment $\Gamma_i(t)$ without assuming that it is negligible.

\textbf{Directional scaling.}
Following the timestep normalization motivated by Eq.~(\ref{eq:separable}), we compute
\begin{equation}
\bar A_i^{\mathrm{eff}}
=\E_t\left[\frac{A_i^{\mathrm{eff}}(t)}{\alpha_t}\right],
\qquad
\bar C_i^{\mathrm{eff}}
=\E_t\left[\frac{C_i^{\mathrm{eff}}(t)}{\sigma_t}\right],
\end{equation}
using the training logit-normal timestep distribution, restricted to the reliable evaluation bins.

\begin{figure}[t]
\centering
\begin{subfigure}[t]{0.32\linewidth}\centering
\includegraphics[width=\linewidth]{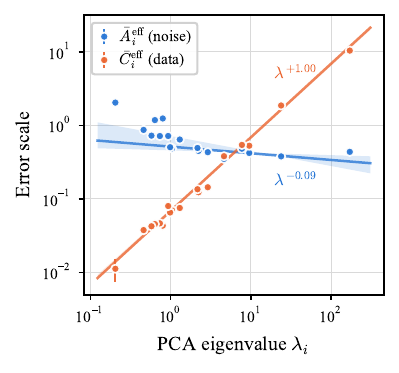}
\caption{Effective error scales}\label{fig:error-validation-a}
\end{subfigure}
\hfill
\begin{subfigure}[t]{0.32\linewidth}\centering
\includegraphics[width=\linewidth]{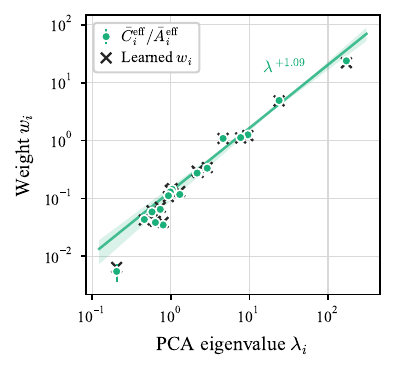}
\caption{Error ratio and learned weights}\label{fig:error-validation-b}
\end{subfigure}
\hfill
\begin{subfigure}[t]{0.32\linewidth}\centering
\includegraphics[width=\linewidth]{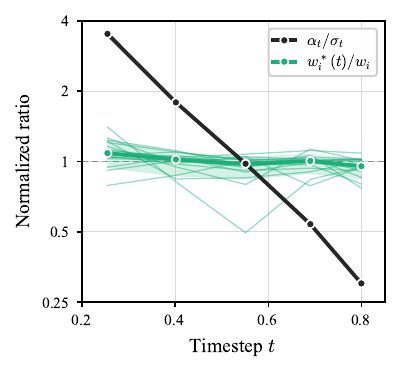}
\caption{Timestep cancellation}\label{fig:error-validation-c}
\end{subfigure}
\caption{\textbf{Estimated prediction-error structure on ImageNet.}
\textbf{(a)} Time-averaged effective noise and data error scales versus PCA variance, with fitted scaling exponents.
\textbf{(b)} Preferred weights estimated from the effective errors compared with the learned AnisoFlow weights.
\textbf{(c)} Timestep dependence of the estimated preferred weights, normalized by the learned weights, compared with the schedule factor $\alpha_t/\sigma_t$ (normalized by its geometric mean). Thin curves show individual directions, and the curve with markers shows the median.
Results are shown for 18 PCA directions satisfying reconstruction-consistency criteria.}
\label{fig:error-validation}
\end{figure}

As shown in Fig.~\ref{fig:error-validation}(a), the effective noise-error scale varies weakly with directional variance, approximately $\bar A_i^{\mathrm{eff}}\propto\lambda_i^{-0.09}$, while the effective data-error scale follows $\bar C_i^{\mathrm{eff}}\propto\lambda_i^{1.00}$. Their ratio scales as $\lambda_i^{1.09}$, closely matching the exponent of $1.07$ for the weights learned by AnisoFlow, fitted over the same 18 directions.

Figure~\ref{fig:error-validation}(b) compares the ratio
$\bar C_i^{\mathrm{eff}}/\bar A_i^{\mathrm{eff}}$ with the learned weights.
The two profiles agree closely in both scaling and magnitude, with a
median ratio $(\bar C_i^{\mathrm{eff}}/\bar A_i^{\mathrm{eff}})/w_i$ of approximately $0.97$.

\textbf{Robustness to direction selection.}
We also relax the reconstruction-consistency threshold from $5\%$ to $10$--$20\%$, increasing the number of analyzed directions from 18 to 45--51 and extending the variance range toward lower-variance directions. The effective-error ratio consistently scales as approximately $\lambda_i^{1.09}$, closely matching the learned-weight exponent of approximately $1.07$ across all thresholds.

\textbf{Timestep dependence.}
We further examine the timestep-dependent preferred weights from Eq.~(\ref{eq:effective-preferred-weight}). Across five reliable timestep bins spanning approximately $t=0.25$ to $0.80$, the explicit schedule factor $\alpha_t/\sigma_t$ varies by $11.7\times$. However, the preferred weights estimated from these effective errors vary by only $1.19\times$ at the median across directions, as shown in Fig.~\ref{fig:error-validation}(c). This indicates that the timestep dependence of the error ratio largely cancels the schedule factor, supporting the use of a single weight profile shared across timesteps.

\subsection{Learning the Basis}
\label{app:basis}

We learn the basis $\mB$ with one of two optimizers: Stiefel Adam, which optimizes $\mB$ directly on the manifold of orthonormal matrices, or the spectral update, a closed-form eigendecomposition step. We describe each in turn.

\subsubsection{Stiefel Adam}
\label{app:stiefel}
Stiefel Adam treats $\mB$ as a parameter constrained to the Stiefel manifold $\{\mB:\mB\T\mB=\mI\}$ and adapts Adam \citep{kingma2014adam} to it. Let $\mathcal P_{\mB}(\mA)=\mA-\mB\,\mathrm{sym}(\mB\T\mA)$, with $\mathrm{sym}(\mA)=\tfrac12(\mA+\mA\T)$, denote the projection onto the tangent space at $\mB$ under the Euclidean metric, and let $\mP$ and $v$ denote the first-moment matrix and a scalar second moment, with step size $\tau$. Each step computes the projected gradient $\bm{\Xi}$ and the updates
\begin{align}
\bm{\Xi}&\leftarrow\mathcal P_{\mB}(\nabla_{\mB}\mathcal L),\qquad
\mP\leftarrow\beta_1\mP+(1-\beta_1)\,\bm{\Xi},\qquad
v\leftarrow\beta_2 v+(1-\beta_2)\,\|\bm{\Xi}\|_F^2/D^2,\nonumber\\
\mB&\leftarrow\mQ,\quad\text{where }\mQ\mR=\mB-\tau\,\hat{\mP}\big/\big(\sqrt{\hat v}+\delta\big),\qquad
\mP\leftarrow\mathcal P_{\mB}(\mP),
\end{align}
where $\hat{\mP}$ and $\hat v$ are the bias-corrected moments, $\delta>0$ is Adam's numerical-stability constant (written $\delta$ rather than $\epsilon$ to avoid a clash with the noise $\bm{\epsilon}$), $\mQ\mR$ denotes the reduced QR decomposition with $\mR$ upper triangular with positive diagonal so that $\mQ$ is unique, and the final projection transports the momentum to the tangent space at the new point. The second moment is a single scalar per basis, so the tangent momentum is scaled by one adaptive step size. An elementwise rescaling would generally leave the tangent space. Weight decay is zero: $\|\mB\|_F^2=D$ is constant on the manifold, and the tangent projection removes the gradient of an $L_2$ penalty. We use the same learning rate and Adam coefficients as for the model parameters (Appendix~\ref{app:impl}). This is a projection-transport Stiefel optimizer under the Euclidean metric. It does not implement parallel transport or the canonical metric. It makes no structural assumption about the loss and serves as the general-purpose alternative to the closed-form update derived next.

\subsubsection{Spectral Update}
\label{app:spectral}
We now derive the spectral update used in our main experiments. Let $\vr := \operatorname{sg}(\hat{\vu}_W) + \vx$, where $\operatorname{sg}$ denotes stop-gradient: the network prediction is held fixed while the basis is updated. In direction $\vb_i$, the recovered-velocity error of Eq.~(\ref{eq:recovery}) is the target-prediction error divided by the recovery denominator, $\hat u_i-u_i=(\vb_i\T\vr-w_i\vb_i\T\bm{\epsilon})/(\alpha_t w_i+\sigma_t)$. Squaring, taking the expectation over $\vx$, $\bm{\epsilon}$, and $t$, and dropping the $w_i^2\vb_i\T\bm{\epsilon}\bm{\epsilon}\T\vb_i$ term, whose contribution $\E_t[w_i^2/(\alpha_t w_i+\sigma_t)^2]$ does not depend on $\vb_i$, the basis-dependent part of the loss is
\begin{equation}\label{eq:basis-objective}
\mathcal L_{\mB}(\mB)=\sum_{i=1}^D\vb_i\T\mM_i\vb_i+\text{const},
\qquad
\mM_i=\E_t\!\left[\frac{\mS_t-2w_i\mH_t}{(\alpha_t w_i+\sigma_t)^2}\right],
\end{equation}
where $\mS_t:=\E[\vr\vr\T\mid t]$ and $\mH_t:=\tfrac12\E[\vr\bm{\epsilon}\T+\bm{\epsilon}\vr\T\mid t]$ are the second moment of $\vr$ and its symmetric cross moment with the noise. This expression is exact given the network output and the weights.

Minimizing Eq.~(\ref{eq:basis-objective}) over orthonormal $\mB$ is not an
eigenvalue problem in general: each column $\vb_i$ is scored by a different
matrix $\mM_i$, because $w_i$ enters both the numerator and the recovery
denominator, so no single eigendecomposition solves it. Direct
optimization on the Stiefel manifold (Appendix~\ref{app:stiefel}) is the
general-purpose alternative. Under the following assumption, however, one
eigendecomposition suffices.
\begin{assumption}\label{ass:joint-diag}
The matrices $\mM_1,\dots,\mM_D$ commute pairwise. Equivalently, they are
simultaneously orthogonally diagonalizable, i.e.,
$\mM_i=\mB_\star\mLambda_i\mB_\star\T$ for a common orthonormal $\mB_\star$ and
diagonal $\mLambda_i=\operatorname{diag}(\kappa_{i1},\dots,\kappa_{iD})$.
\end{assumption}
\begin{proposition}[Optimal basis under joint diagonalizability]\label{prop:spectral}
Under Assumption~\ref{ass:joint-diag}, a global minimizer of
Eq.~(\ref{eq:basis-objective}) over $\mB\T\mB=\mI$ is
$\mB=\mB_\star\bm{\Pi}\operatorname{diag}(s_1,\dots,s_D)$ for some permutation
matrix $\bm{\Pi}$ and signs $s_i\in\{\pm1\}$. Moreover, for any coefficients
$\gamma_i$ such that $\sum_i\gamma_i\mM_i$ has distinct eigenvalues, the
eigenvectors of $\sum_i\gamma_i\mM_i$ recover the common eigendirections.
An optimal pairing assigns column $i$ to direction $\pi(i)$ minimizing
$\sum_i\kappa_{i,\pi(i)}$ over permutations $\pi$.
\end{proposition}
\begin{proof}
Write $\mB=\mB_\star\bm{\Omega}$ with $\bm{\Omega}$ orthogonal. Then
$\vb_i\T\mM_i\vb_i=\sum_j\kappa_{ij}\Omega_{ji}^2$, so
$\mathcal L_{\mB}=\sum_{i,j}\kappa_{ij}\Omega_{ji}^2+\text{const}$ is linear in the
matrix $[\Omega_{ji}^2]$, which is doubly stochastic because $\bm{\Omega}$ is orthogonal. A
linear function on the set of doubly stochastic matrices attains its minimum at
a permutation matrix (Birkhoff--von Neumann), and every permutation matrix is
itself an orthogonal $\bm{\Omega}$, so the minimum is attained at
$\bm{\Omega}=\bm{\Pi}\operatorname{diag}(s_1,\dots,s_D)$. The second claim follows
because $\sum_i\gamma_i\mM_i=\mB_\star(\sum_i\gamma_i\mLambda_i)\mB_\star\T$ has
eigenvectors $\mB_\star$ whenever its eigenvalues are distinct.
\end{proof}
We use the uniform average, which weights every output direction once, as in
the squared loss, and introduces no directional preference. Substituting
Eq.~(\ref{eq:basis-objective}),
\begin{equation}\label{eq:basis-moment}
\mG:=\frac1D\sum_{i=1}^D\mM_i
=\E_t\big[a_t\,\mS_t-2b_t\,\mH_t\big]
=\E\big[a_t\,\vr\vr\T-b_t(\vr\bm{\epsilon}\T+\bm{\epsilon}\vr\T)\big],
\end{equation}
with
\begin{equation}
    a_t
    :=
    \frac{1}{D}\sum_{j=1}^D
    \frac{1}{(\alpha_t w_j+\sigma_t)^2},
    \qquad
    b_t
    :=
    \frac{1}{D}\sum_{j=1}^D
    \frac{w_j}{(\alpha_t w_j+\sigma_t)^2}.
\end{equation}
The spectral update sets $\mB\leftarrow\operatorname{eigvecs}(\mG)$.
Under Assumption~\ref{ass:joint-diag}, these are the common eigendirections
whenever the eigenvalues of $\mG$ are distinct. Commutativity alone does not
guarantee this: a repeated eigenvalue of $\mG$ determines the corresponding
eigenvectors only up to a rotation within their span. The cross term accounts for the correlation between
$\vr$ and the noise in the target, and omitting it changes the basis objective.

\begin{remark}[Pairing eigenvectors with weights]\label{rem:pairing}
We order the eigenvectors of $\mG$ by ascending eigenvalue and keep $w_i$ attached to rank $i$. This rank-based pairing is a heuristic; it need not solve the assignment problem of Proposition~\ref{prop:spectral}, but it keeps each weight attached to a stable direction as long as eigenvalue ranks do not cross. Adam continues to adapt the weights after each basis update. Column signs cancel in $\mB\operatorname{diag}(w_i)\mB\T$.
\end{remark}

When Assumption~\ref{ass:joint-diag} holds only approximately,
$\mM_i=\mB_\star\mLambda_i\mB_\star\T+\bm{\Delta}_i$ with small non-commuting
residuals $\bm{\Delta}_i$, we have $\mG=\mB_\star\bar{\mLambda}\mB_\star\T+\bar{\bm{\Delta}}$
with $\bar{\mLambda}=\frac1D\sum_i\mLambda_i$ and $\bar{\bm{\Delta}}=\frac1D\sum_i\bm{\Delta}_i$, and standard eigenvector perturbation
bounds keep $\operatorname{eigvecs}(\mG)$ close to $\mB_\star$ whenever
$\|\bar{\bm{\Delta}}\|$ is small relative to the eigengaps of $\bar{\mLambda}$. The
spectral update is then an approximate joint-diagonalization step.
Appendix~\ref{app:linear-fixed-point} shows that the assumption holds exactly
at covariance-aligned solutions in a linear-network model.

\textbf{Online estimator.}
During training we estimate $\mG$ from minibatches. For a batch of $M$ images
with $N$ patches each, the estimate is
\begin{equation}
\widehat{\mG}=\frac{1}{MN}\sum_{m=1}^M\sum_{p=1}^N
\big[a_{t_m}\vr_{mp}\vr_{mp}\T-b_{t_m}(\vr_{mp}\bm{\epsilon}_{mp}\T+\bm{\epsilon}_{mp}\vr_{mp}\T)\big],
\end{equation}
normalized by the number of patches rather than by the random sum of recovery
weights, so that it estimates a fixed population quantity. We maintain an
exponential moving average $\mG\leftarrow(1-\beta)\mG+\beta\,\widehat{\mG}$ and
set $\mB$ to the eigenvectors of $\mG$ at every optimization step
(Appendix~\ref{app:impl}). The moving average smooths the moment estimate, although individual
eigenvectors can still change abruptly near eigenvalue crossings.

\textbf{Relation to PCA initialization.} When the network output is zero, as is common at initialization, $\vr=\vx$. The data and sampled noise are independent, so $\E[\vx\bm{\epsilon}\T]=\vzero$, and $a_t$ depends only on the timestep and the current weights. Eq.~(\ref{eq:basis-moment}) therefore reduces in expectation to a scalar multiple of the data second moment, $\mG\propto\E[\vx\vx\T]$, which for centered data is the covariance, so its eigenvectors are the PCA directions. In practice the running moment is initialized to $\mI$ (Appendix~\ref{app:impl}); since adding a multiple of the identity leaves eigenvectors unchanged, the first basis consists of the eigendirections of a single minibatch estimate, a noisy PCA basis rather than the exact one. The first spectral update from $\mW=\mI$ thus recovers PCA structure on its own, which is why PCA initialization is optional. Appendix~\ref{app:linear-fixed-point} studies when this alignment is preserved as training proceeds.

\subsection{Linear-Model Analysis of Basis Alignment}
\label{app:linear-fixed-point}

We analyze the spectral update in an idealized setting: the weights are fixed, the
network is replaced by the optimal time-conditioned linear predictor, and the basis is
updated by the exact iteration $\mB\leftarrow\operatorname{eigvecs}(\mG)$ in place of the
moving-average estimate of Appendix~\ref{app:spectral}. We show that this iteration
preserves alignment with the data covariance and give a condition under which the
alignment is locally stable.

\textbf{Linear denoising model.}
Let $\mSigma=\E[\vx\vx\T]=\mV\operatorname{diag}(\lambda_1,\dots,\lambda_D)\mV\T$
be the covariance of centered data, and let
$\mSigma_{z,t}=\E[\vz_t\vz_t\T]=\alpha_t^2\mSigma+\sigma_t^2\mI$.
For a symmetric $\mW$ and a linear predictor $\hat{\vu}_W=\mPhi_t\vz_t$, minimizing the
recovered-velocity loss over $\mPhi_t$ gives the same minimizer as least squares on the
target, since recovery applies the fixed invertible map $(\alpha_t\mW+\sigma_t\mI)^{-1}$
to the target error. Hence
\begin{equation}\label{eq:linear-predictor}
\mPhi_t^*(\mW)
=(\sigma_t\mW-\alpha_t\mSigma)\mSigma_{z,t}^{-1},
\end{equation}
since $\E[\vu_W\vz_t\T]=\sigma_t\mW-\alpha_t\mSigma$.
Substituting into the recovery map of Eq.~(\ref{eq:recovery}) gives
\begin{equation}
(\alpha_t\mW+\sigma_t\mI)^{-1}
(\mI-\mW+\mPhi_t^*)
=(\sigma_t\mI-\alpha_t\mSigma)\mSigma_{z,t}^{-1}.
\end{equation}
Thus the recovered velocity predictor, and hence the optimal linear loss, do
not depend on $\mW$. The linear model therefore identifies neither the weights
nor the basis, and the alignment shown below is a property of the spectral
update itself rather than of the loss.

\textbf{Preservation of covariance alignment.}
For the fitted predictor, the moments in Eq.~(\ref{eq:basis-objective}) are
\begin{align}
\mS_t&=\sigma_t^2\big(\mW\mSigma_{z,t}^{-1}\mW+\mSigma\mSigma_{z,t}^{-1}\big),\\
\mH_t&=\frac{\sigma_t^2}{2}
\big(\mW\mSigma_{z,t}^{-1}+\mSigma_{z,t}^{-1}\mW\big)
-\alpha_t\sigma_t\mSigma\mSigma_{z,t}^{-1}.
\end{align}
These formulas follow by substituting
$\vr=(\alpha_t\mPhi_t^*+\mI)\vx+\sigma_t\mPhi_t^*\bm{\epsilon}$, using
$\alpha_t\mPhi_t^*+\mI=\sigma_t(\alpha_t\mW+\sigma_t\mI)\mSigma_{z,t}^{-1}$, the
independence of $\vx$ and $\bm{\epsilon}$, and the fact that $\mSigma_{z,t}$ commutes
with $\mSigma$. If $\mW$ commutes with $\mSigma$, all of these
matrices, and hence every $\mM_i$ and $\mG$, are diagonal in a common eigenbasis of
$\mW$ and $\mSigma$, so
Assumption~\ref{ass:joint-diag} holds exactly. When $\mG$ has distinct
eigenvalues, the update returns the PCA basis up to sign and permutation, so
alignment with the data covariance is preserved. If the rank-based pairing of
Remark~\ref{rem:pairing} also leaves each weight attached to its direction,
$\mW$ itself is a fixed point. When $\mG$ has a repeated eigenvalue, the update may rotate within the tied eigenspace. Arbitrary rotations within that space are guaranteed to preserve alignment if the restriction of $\mSigma$ to that eigenspace is a multiple of the identity.

For a nearly aligned $\mW$, the identity
\begin{equation}
[\mPhi_t^*,\mSigma]
=\sigma_t[\mW,\mSigma]\mSigma_{z,t}^{-1}
\end{equation}
shows that a small misalignment of $\mW$ induces a proportionally small
misalignment of the predictor. Since the moments are continuous in $\mW$, the
eigenvectors of $\mG$ also vary continuously as long as its eigenvalues stay
separated.

\textbf{Local stability.}
Consider an aligned fixed point $\mW_\star=\operatorname{diag}(w_1,\dots,w_D)$
in the PCA basis whose rank-based pairing (Remark~\ref{rem:pairing}) is locally
unchanged, and assume the eigenvalues $\bar\kappa_j=\frac1D\sum_i\kappa_{ij}$ of
$\mG$ there, in the notation of Appendix~\ref{app:spectral}, are distinct.
Set $d_j(t)=\alpha_t^2\lambda_j+\sigma_t^2$, the eigenvalues of $\mSigma_{z,t}$.
Differentiating the moment
formulas gives, for an off-diagonal perturbation,
\begin{equation}
\delta G_{jk}=L_{jk}\,\delta W_{jk},\qquad
L_{jk}=\E_t\!\left[\sigma_t^2\left\{
 a_t\left(\frac{w_j}{d_j(t)}+\frac{w_k}{d_k(t)}\right)
-b_t\left(\frac{1}{d_j(t)}+\frac{1}{d_k(t)}\right)
\right\}\right].
\end{equation}
A small rotation of the basis by an angle $\theta_{jk}$ in the $(j,k)$ plane
gives $\delta W_{jk}=(w_j-w_k)\theta_{jk}$, so
\begin{equation}
\theta_{jk}^{\mathrm{new}}
=\rho_{jk}\theta_{jk}+O(\|\bm{\theta}\|^2),\qquad
\rho_{jk}=\frac{L_{jk}(w_j-w_k)}{\bar\kappa_j-\bar\kappa_k}.
\end{equation}
The fixed point is locally attractive when $|\rho_{jk}|<1$ for every pair
with $w_j\neq w_k$ (rotations between equal weights leave $\mW$ unchanged).
Larger eigengaps of $\mG$ reduce sensitivity for a fixed numerator in $\rho_{jk}$. At an exact tie, the corresponding eigenvectors are not determined by $\mG$.

\textbf{Scope.} These results concern the idealized linear iteration. They do not establish global convergence, nor stability of the joint dynamics of the network, the weights, and the running moment. Empirically, the basis learned with a nonlinear network aligns closely with the PCA basis (Fig.~\ref{fig:learned-w-b}).

\section{Implementation Details}
\label{app:impl}

\subsection{AnisoFlow Parameterization and Optimization}
\label{app:param}

\textbf{Learning the directional weights.}
We write
\begin{equation}
\mW = \mB\operatorname{diag}(w_1,\ldots,w_D)\mB\T,
\qquad w_i\geq0,
\end{equation}
and learn the directional weights jointly with the network. To enforce
strict positivity, we set $w_i=\exp(\eta_i)$ and optimize $\eta$ with Adam. Unless otherwise noted, $\eta$ uses the same learning rate as the model parameters.
By default, the weights are shared across timesteps
and are optimized from the start of training.

\textbf{Learning the basis.}
By default, the orthogonal basis $\mB$ is updated during training using the
spectral update of Appendix~\ref{app:basis}. The running moment
$\mG$ is computed from the stop-gradient network prediction. It is
initialized to $\mI$ and updated with an exponential moving average with
coefficient $\beta=10^{-5}$. At every optimization step, the running moment is
averaged across devices, eigendecomposed, and $\mB$ is set to its eigenvectors
ordered by ascending eigenvalue, with $w_i$ attached to eigenvalue rank $i$.
Alternatively, Stiefel Adam (Appendix~\ref{app:stiefel}) optimizes $\mB$ directly with the same learning rate and Adam coefficients as the model parameters; in this variant the basis of the EMA model is re-orthogonalized by QR
after each EMA step.

\textbf{Numerical details.}
At inference, the recovery denominator $\alpha_t w_i+\sigma_t$ is
lower-bounded by $\sigma_{\min}=4\times10^{-2}$, following AsymFlow
\citep{asymflow} for the baselines and AnisoFlow alike.

\subsection{ImageNet Pixel-Space Experiments}
\label{app:imagenet}

\textbf{Training setup.}
We conduct our main pixel-space experiments on ImageNet $256{\times}256$ using
the same architecture, optimizer, and training hyperparameters as JiT-H/16
(see Table~9 of JiT~\citep{jit}), changing only the prediction-target
parameterization. We use JiT-B/16 for smaller-scale experiments. Unless
otherwise stated, the main models are trained for 600 epochs.

\textbf{Data covariance and PCA estimation.}
The $\mW=\mI$ AnisoFlow variants do not require a PCA basis or a precomputed
data covariance. We estimate the covariance for the PCA-initialized and
fixed-PCA experiments and for analyzing the learned directions and weights. For JiT with $16{\times}16$
patches, each RGB patch is represented as a
$16{\times}16{\times}3=768$-dimensional vector. We compute the empirical
covariance of mean-centered patches from $100{,}000$ training images, in the
same normalized pixel space seen by the network, and eigendecompose it as
$\mSigma=\mV\operatorname{diag}(\lambda_1,\ldots,\lambda_D)\mV^\top$.
Here, $\mV$ denotes the PCA basis and $\lambda_i$ the corresponding PCA
eigenvalues used in the analyses of \S\ref{sec:theory}.
For JiT-H/16, $\mW\in\mathbb{R}^{768\times768}$ acts independently on each
image patch and is shared across spatial locations; for an image containing
$N$ patches, the corresponding image-level operator is
$\mI_N\otimes\mW$.

\textbf{Classifier-free guidance.}
Following AsymFlow \citep{asymflow}, for each AnisoFlow model, we grid-search the CFG scale \citep{ho2021cfg} with step size $0.1$ and the guidance interval with step size $0.02$.
Table~\ref{tab:imagenet-cfg} lists the selected settings for
Table~\ref{tab:target-ablation}.

\begin{table}[H]
\centering
\caption{\textbf{Selected guidance settings for the AnisoFlow models in
Table~\ref{tab:target-ablation}.} CFG scale / guidance interval selected by grid
search for each model and training length.}
\label{tab:imagenet-cfg}
\scalebox{0.85}{%
\begin{tabular}{llcc}
\toprule
\textbf{Noise basis} & \textbf{Init} & \textbf{@160ep} & \textbf{@600ep} \\
\midrule
Frozen (PCA)    & PCA & $3.1$ / $[0,0.76]$ & $2.3$ / $[0,0.88]$ \\
Stiefel Adam    & PCA & $3.1$ / $[0,0.74]$ & $2.1$ / $[0,0.90]$ \\
Spectral update & PCA & $3.1$ / $[0,0.74]$ & $2.3$ / $[0,0.88]$ \\
Stiefel Adam    & --  & $4.4$ / $[0,0.62]$ & $2.3$ / $[0,0.84]$ \\
Spectral update & --  & $3.6$ / $[0,0.66]$ & $2.1$ / $[0,0.86]$ \\
\bottomrule
\end{tabular}
}
\end{table}

\subsection{Text-to-Image Experiments}
\label{app:t2i}

We evaluate AnisoFlow on pixel-space text-to-image generation with
JiT-L/16-T2I as the baseline, which uses the MM-JiT denoiser architecture from
MiniT2I~\citep{minit2i} and the T5Gemma~\citep{t5gemma} text encoder from i1~\citep{zeng2026i1}.
The baseline uses clean-data prediction, and AnisoFlow
differs from it only in the prediction target. Both models are trained on i1's data~\citep{zeng2026i1}, which combines the public ImageNet-22K~\citep{deng2009imagenet}, Places365~\citep{places365}, Pexels~\citep{pexels}, TextAtlas~\citep{textatlas}, RedCaps~\citep{redcaps}, and Megalith-10M~\citep{megalith10m} with i1's rendered-text and synthetic sources, plus YFCC100M~\citep{yfcc100m} and the released iNaturalist split~\citep{inaturalist} in the $256{\times}256$ stage only. We follow an i1-equivalent schedule through the $256{\times}256$ and $512{\times}512$ stages, doubling the batch size to $1024$ while halving the number of iterations. As in the pixel-space ImageNet models, $\mW\in\mathbb{R}^{768\times768}$ acts
on $16{\times}16{\times}3$ patches. AnisoFlow uses the PCA initialization and
the spectral update (EMA coefficient $10^{-5}$) during the $256{\times}256$
tier. During the $512{\times}512$ tier, the basis is frozen while the weights
$w_i$ continue to be learned. For hyperparameters that cannot be shared between pixel and latent models (e.g., flow shift~\citep{esser2024sd3}), we tune them ourselves. Table~\ref{tab:t2i-hparams} summarizes the settings of each training stage.

\begin{table}[t]
\centering
\caption{\textbf{Text-to-image model and training settings.} The $256{\times}256$ and $512{\times}512$ stages are used for the quantitative comparison, and the one-megapixel (1MP) and alignment fine-tuning (SFT) stages only for the qualitative samples in Fig.~\ref{fig:teaser}. Each stage is initialized from the previous one.}
\label{tab:t2i-hparams}
\resizebox{\linewidth}{!}{%
\begin{tabular}{lcccc}
\toprule
\textbf{Setting} & $\mathbf{256{\times}256}$ & $\mathbf{512{\times}512}$ & \textbf{1MP} & \textbf{SFT} \\
\midrule
Pixel color space & \multicolumn{4}{c}{Normalized Oklab~\citep{ottosson2020oklab,asymflow}} \\
Patch size / dimension $D$ & \multicolumn{4}{c}{16 / 768} \\
Text encoder & \multicolumn{4}{c}{T5Gemma-2B~\citep{t5gemma}, 256 tokens} \\
AnisoFlow weights & \multicolumn{4}{c}{Learned} \\
AnisoFlow basis & PCA init, spectral update & Frozen & Frozen & Frozen \\
\midrule
Optimizer & \multicolumn{4}{c}{Adam~\citep{kingma2014adam}, betas $(0.9, 0.95)$, weight decay $0$} \\
Precision & \multicolumn{4}{c}{BF16 autocast, FP32 master weights} \\
Prompt dropout & \multicolumn{4}{c}{$0.1$} \\
EMA & \multicolumn{4}{c}{Dynamic EMA~\citep{karras2024edm2}, $\gamma=7.0$, reset at each stage} \\
Resolution & $256{\times}256$ & $512{\times}512$ & 1MP, mixed aspect ratios & 1.56MP, mixed aspect ratios \\
Training data & i1 & i1 & i1 (real only) + LAION-3M & Z-Image Turbo synthetic \\
Batch size & 1024 & 1024 & 256 & 256 \\
Training iterations & 1M & 250K & 150K & 2K \\
Learning rate & $2\times10^{-4}$ & $2\times10^{-4}$ & $2\times10^{-4}$ & $10^{-5}$ \\
Warmup iterations & 1000 & -- & -- & 100 \\
Pre-shift time sampling & $\mathrm{LogitNormal}(0,1)$ & Uniform & Uniform & Uniform \\
Flow shift & 2.0 & 3.0 & 5.0 & 6.0 \\
\bottomrule
\end{tabular}
}
\end{table}

\textbf{Evaluation.}
All models are sampled with a 32-step SDE solver based on
UniPC~\citep{zhao2023unipc} with the stochastic coefficient of
ER-SDE~\citep{cui2023ersde}, using the same guidance settings for the quantitative comparison: a guidance scale of 6.0 on the interval $t\in[0,0.99]$ with orthogonal projection from APG~\citep{sadat2025oversaturation}. For benchmarking, we use i1's
rewritten prompts for LongText-Bench, GenEval, and DPG-Bench, while other benchmarks use their original prompts.

\textbf{Reduced-precision inference.}
To evaluate numerical stability, we sample the JiT baseline and AnisoFlow models with the weights and activations of the transformer's linear layers quantized to 8 bits (W8A8), while keeping prompts, seeds, and
sampler settings fixed.

\textbf{High-resolution qualitative samples.}
The one-megapixel samples in Fig.~\ref{fig:teaser} come from an additional
one-megapixel tier of the AnisoFlow model, which we do not evaluate
quantitatively due to the additional cost of training a matching baseline. 
For this tier, we remove the synthetic-data mix from i1's data and add the 3M LAION subset~\citep{schuhmann2022laion5b} used by
AsymFlow~\citep{asymflow}, which has mixed aspect ratios, followed by a short alignment fine-tuning stage on
synthetic images from Z-Image Turbo~\citep{zimage2025}, rendered at about 1.56 megapixels, with a reduced learning rate of $10^{-5}$.

\subsection{Latent-Space Experiments}
\label{app:latent-exp}

\textbf{SiT-XL/2-REPA.}
We evaluate AnisoFlow on SiT-XL/2-REPA, preserving the original latent
representation, model architecture, representation-alignment objective, and
training recipe while changing only the prediction-target parameterization.
The baseline uses velocity prediction. The model operates on SD-VAE latents
\citep{ldm} with patch size $2$, so $\mW\in\mathbb{R}^{16\times16}$ acts on
each $2{\times}2{\times}4$ latent patch. We initialize $\mW=\mI$ and use the
spectral update with EMA coefficient $10^{-5}$.

\textbf{RAEv2.}
We also evaluate AnisoFlow on RAEv2, preserving its representation, architecture, and Muon-based training recipe~\citep{jordan2024muon,GramNewtonSchulz}. We make two changes to the published recipe, applied to both our RAEv2 baseline and Aniso-RAEv2: timesteps are sampled uniformly before the shift rather than logit-normally, and the inference clamp is set to $\sigma_{\min}=0.25$. These changes improve the baseline itself (JiT FID $1.06\to0.98$), so the comparison remains fair. The baseline uses clean-data prediction. AnisoFlow acts on the $1024$-dimensional RAEv2 tokens, so $\mW\in\mathbb{R}^{1024\times1024}$, with the PCA initialization ($w_i=0.15\lambda_i$ in the PCA basis of the tokens). Because RAEv2 trains for only 80 epochs, we raise the spectral-update EMA coefficient to $10^{-4}$ and the noise-weight learning rate to $10\times$ the model learning rate for faster convergence.

\subsection{Compute Overhead}
\label{app:overhead}
AnisoFlow adds $D$ learned directional weights and maintains the learned
orthogonal basis $\mB\in\mathbb{R}^{D\times D}$. During training, the spectral
basis update additionally maintains a $D\times D$ running moment matrix and
eigendecomposes it at every step to update $\mB$. At each model evaluation,
prediction and velocity recovery require transforms into and out of the learned
basis together with elementwise directional operations. The eigendecomposition
of the $D\times D$ running moment is performed at every optimization step during
training only; at inference time, the learned basis is fixed.

The resulting measured training and inference overheads relative to JiT-H/16
are reported in Table~\ref{tab:target-ablation}.

\begin{table}[H]
\centering
\caption{\textbf{Additional computation and state introduced by AnisoFlow.}}
\label{tab:overhead}
\scalebox{0.85}{%
\begin{tabular}{lll}
\toprule
\textbf{Component} & \textbf{Additional state / operation} &
\textbf{When applied} \\
\midrule
Directional weights
    & $D$ scalars
    & Training and inference \\
Learned basis
    & $D\times D$ orthogonal matrix
    & Training and inference \\
Running spectral moment
    & $D\times D$ matrix
    & Training only \\
Basis transforms
    & Multiplication by $\mB$ and $\mB^\top$
    & Each model evaluation \\
Directional recovery
    & Elementwise scaling
    & Each model evaluation \\
Spectral basis update
    & Eigendecomposition of a $D\times D$ matrix
    & Every training step \\
\bottomrule
\end{tabular}
}
\end{table}

\raggedbottom
\section{Additional Experiments}
\label{app:exp-extra}

\begin{figure}[H]\centering
\includegraphics[width=.5\linewidth]{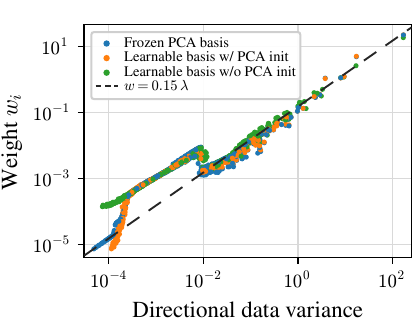}
\caption{\textbf{Effect of the JiT input bottleneck.}
Untrimmed version of Fig.~\ref{fig:learned-w-a}, showing all 768 directional
variance/weight pairs for the frozen PCA basis and learnable bases with and
without PCA initialization. Both learnable variants use spectral updates;
the dashed line shows $w=0.15\lambda$. The near-linear profile at high
variance changes noticeably in the low-variance directions beyond roughly
the 256-dimensional patch-embedding bottleneck.}
\label{fig:bottleneck-spectrum}
\end{figure}

\textbf{Effect of the input bottleneck.}
JiT represents each $16\times16$ RGB patch as a 768-dimensional pixel vector,
which is projected through a 256-dimensional linear patch embedding before
entering the transformer. Thus, only a 256-dimensional projection of the
original patch is directly available to the network. To test whether the
learned target reflects this architectural bottleneck, we learn weights across
the full 768-dimensional PCA spectrum. The learned weights follow the
near-linear spectral profile of the main experiments through roughly the
patch-embedding dimension, but change noticeably beyond this point
(Fig.~\ref{fig:bottleneck-spectrum}). The transition near dimension 256
suggests that the learned target is sensitive to the network's input bottleneck.
The zero-skill limit of Appendix~\ref{app:learned-weights} provides a limiting
case in which directional information is entirely unavailable.

\begin{figure}[H]\centering
\begin{subfigure}[t]{0.38\linewidth}\centering
\includegraphics[width=\linewidth]{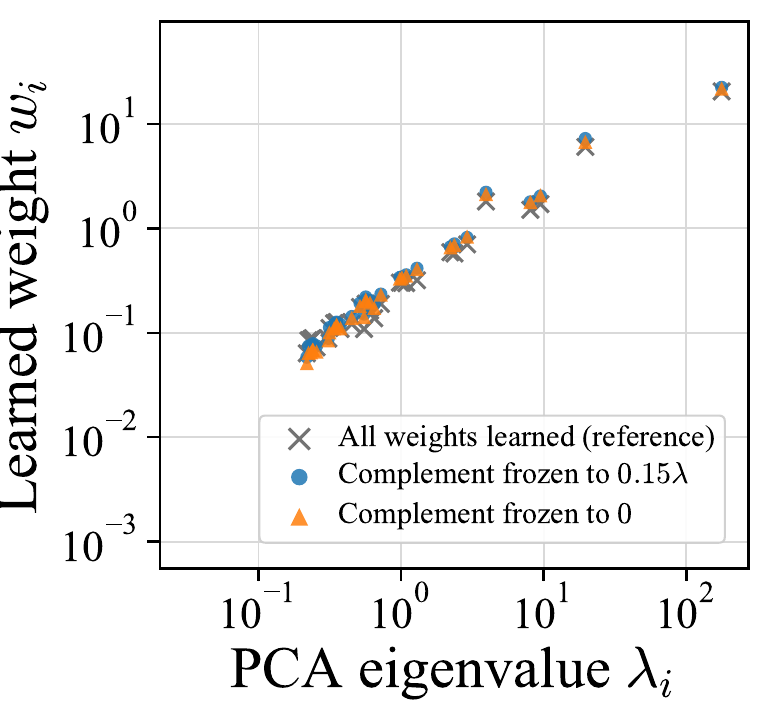}
\caption{Top-32 directions free}\label{fig:coupling-a}
\end{subfigure}
\hspace{0.02\linewidth}
\begin{subfigure}[t]{0.38\linewidth}\centering
\includegraphics[width=\linewidth]{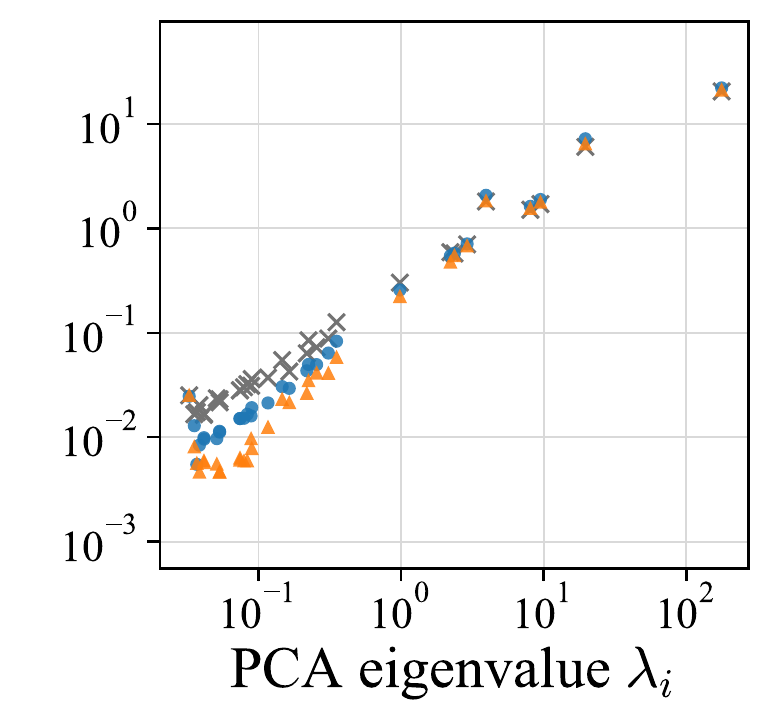}
\caption{Stratified directions free}\label{fig:coupling-b}
\end{subfigure}
\caption{\textbf{Cross-direction coupling.}
We leave 32 directional weights free to learn while freezing all complementary
directions to either $w=0$ or $w=0.15\lambda$. Each point shows the
final learned weight of one of these 32 free directions; gray crosses show the
corresponding weights from a reference model in which all directions are
learned jointly. \textbf{(a)}~The free subset contains the 32 highest-variance PCA
directions. \textbf{(b)}~The free subset is stratified across the data spectrum.
The high-variance free weights closely match the fully learned solution under
both frozen backgrounds, while the lowest-variance directions show somewhat
greater sensitivity to the complementary targets.}
\label{fig:coupling}
\end{figure}

\textbf{Cross-direction coupling.}\label{app:cross-coupling}
Although our theory analyzes each direction separately, the shared network can
couple the preferred weights across directions. Using JiT-B/16 with a fixed PCA
basis, we select 32 directions whose weights remain learnable and freeze all
other directional weights to a fixed background profile. As shown in Fig.~\ref{fig:coupling}, for the highest-variance directions, the free weights remain close to the
fully learned solution under both frozen backgrounds, whereas lower-variance directions exhibit somewhat
greater sensitivity to the complementary targets. These results support the
direction-wise approximation used in our analysis while indicating mild
cross-direction coupling in the lowest-variance directions.

\end{document}